%% file: Master_document.tex
\documentclass[article]{elsarticle}

\journal{Journal of Information Processing and Management}

\usepackage{floatrow}
\usepackage[numbers]{natbib}
\usepackage{lscape}
\usepackage{graphicx}
\usepackage{cite}
\usepackage[lined,ruled,linesnumbered]{algorithm2e}
\usepackage{relsize}
\usepackage{algorithmicx,algpseudocode}
\usepackage{ mathrsfs }
\usepackage{amsmath, amsfonts}
\usepackage{booktabs} % \toprule, \midrule, \bottomrule
\usepackage{array}    % \newcolumntype
\usepackage[below]{placeins} % use \FloatBarrier in the body
\usepackage{caption} %many figures in one figure (note subfigure and subfig are eprecated)
\usepackage{subcaption} %many figures in one figure (note subfigure and subfig are deprecated)
\usepackage{relsize}
\newfloatcommand{capbtabbox}{table}[][\FBwidth]

\newtheorem{pro}[subsection]{Proposition}

\newtheorem{rem}[subsection]{Remark}
\newtheorem{pr}[subsection]{proof}

\begin{document}

\begin{frontmatter}

%++++++++++++++++++++++++++
%\author{\uppercase{Emmanuel Tuyishimire}\authorrefmark{1},% \IEEEmembership{Fellow, IEEE},
%\uppercase{Bigomokero Antoine Bagula\authorrefmark{2}%, and Third C. Author,
%%Jr}.\authorrefmark{3},
%%\IEEEmembership{Member, IEEE}
%}
%}
%\address[1]{University of the Western Cape, South Africa, Cape Town (temmanuel@uwc.ac.za)}
%\address[1]{Department of Computer Science}
%
%%\tfootnote{This paragraph of the first footnote will contain support 
%%information, including sponsor and financial support acknowledgment. For 
%%example, ``This work was supported in part by the U.S. Department of 
%%Commerce under Grant BS123456.''}
%
%\markboth
%{Emmanuel Tuyishimire \headeretal: Preparation of Papers for IEEE TRANSACTIONS and JOURNALS}
%{Bigomokero Antoine Bagula \headeretal: Preparation of Papers for IEEE TRANSACTIONS and JOURNALS}
%
%\corresp{Corresponding author: Emmanuel Tuyishimire (temmanuel@uwc.ac.za)}
%
%

%++++++++++++++++++++++

\title{Real-Time  Data Muling  using a team of heterogeneous unmanned aerial vehicles}
%\tnotetext[mytitlenote]{This is a footnote associated with the title.}
%\tnotetext[mytitlenote2]{Here we go.}
\author{Emmanuel Tuyishimire$^1$, Antoine Bagula$^1$, Slim Rekhis$^2$, Noureddine Boudriga$^2$ }
\address{$^1$ Computer Science Department, University of the Western Cape, Cape Town, South Africa}
\address{$^2$ Communication Networks and Security Research Laboratory, Cartage University, Tunis, Tunisia}

%\author{$^1$temmanuel@uwc.ac.za}
%\corresp{Corresponding author: Emmanuel Tuyishimire (temmanuel@uwc.ac.za)}

%\cortext[mycorrespondingauthor]{This is to indicate the corresponding author.}


\begin{abstract}
The use of Unmanned Aerial Vehicles (UAVs) in Data transport has attracted a lot of attention and applications, as a modern traffic engineering technique used in data sensing, transport, and delivery to where infrastructure is available for its interpretation. Due to UAVs constraints such as limited power lifetime, it has been necessary to assist them with ground sensors to gather local data which has to be transferred to UAVs upon visiting the sensors. The management of such ground sensor communication together with a team of flying UAVs constitutes an interesting  data muling problem which still deserves to be addressed and investigated. This paper revisits the issue of traffic engineering in Internet-of-Things (IoT) settings, to assess the relevance of using UAVs for the persistent collection of sensor readings from the sensor nodes located into an environment and their delivery to base stations where further processing is performed. We propose a persistent path planning and UAV allocation model, where a team of heterogeneous UAVs coming from various  base stations are used to collect data from ground sensors and deliver the collected information to their closest base stations. This problem is mathematically formalised as a real-time constrained optimisation model, and proven to be NP-hard. The paper proposes a heuristic solution to the problem and evaluates its relative efficiency through performing experiments on both artificial and real sensors networks, using various scenarios of UAVs settings.
\end{abstract}

\begin{keyword}
Real-time visitation, cooperative UAVs, path planing, clustered network.
\end{keyword}

\end{frontmatter}

%\linenumbers

\input{multidrone}

\bibliographystyle{unsrt}
\bibliography{references}
\addcontentsline{toc}{chapter}{References}

%\section*{References}

%\bibliography{mybibfile}

\end{document}

%% file: multidrone.tex
\section{Introduction}

The use of UAVs has  emerged as a flexible and cost-efficient alternative to traditional traffic engineering techniques which have been used  in  IoT settings, to transport sensor readings from their points of collection to their processing places.  However, the joint path finding and resource allocation for a team when tasked to achieve collaborative data muling, is still an issue that require further investigations. On the other hand, while accurate solutions to data muling problems are still scarce, especially when considering the limited flying capacity of the battery-powered UAVs, issues related to the efficient task allocation to a team of UAVs under stringent data collection requirements, such as real time data collection, still need to be addressed, especially when UAVs have different specifications (speeds, battery, lifetime, memory, functionalities, etc) and only flesh and complete information need to be collected. Furthermore, persistent collection requirement needs to be addressed and this requires the data muling system to deal with outdated or premature  sensor readings. 

Potential applications of a such real-time data muling model include (i) city surveillance in order to evaluate risks and respond with appropriate actions by having a team  of UAVs persistently visiting locations of interests in a smart city for public safety, parking spots localization \citep{smartparking} and pollution 
monitoring \citep{pollution} ; (ii) drought mitigation to support small scale farming in rural areas \citep{masinde1,masinde2} by using a team of UAVs to 
collect farmland image collection and processing these images to achieve situation recognition for precision irrigation; (iii) periodic surveillance of buildings and cities' infrastructures for structural health monitoring and maintenance; and (iv) extension of the reach of community mesh networks in rural settings for healthcare \citep{healthcare1,healthcare2} by using a team of UAVs (such as drones) as wireless access points. 
 
%The use of UAVs for visiting ground sensors has been a subject of research.
Sensors  visitation under the fuel consumption constraints was addressed in  \citep{las2012persistent}, and  the visitation under the revisit deadline constraint was proposed in \citep{las2013persistent}. Both works assumed a  single moving agent (UAV) which  optimally visits  various targets. \citep{las2015cooperative} proposes  a cooperative UAVs model where many targets  are visited by a  team of UAVs for persistent  surveillance and pursuit. In this work, the UAVs do not communicate with each other but rather rely on  the information from the  static underground sensors, which are  optimally placed as proposed in \citep{las2015optimal}. However,
all these models do not consider the persistent data delivery and heterogeneity of UAVs which might have different fabrics and  characteristics. Furthermore, neither the energy/battery consumption while the UAVs are waiting  for  the updated  information from the terrestrial sensor network nor the penalty associated with stale information due to late visitation by the  UAV to the sensor nodes have been accounted for. While models were proposed in \citep{tuyishimireinternet, bagulainternet,bender2015minimum, citovsky2015exact}  for the periodic and persistent UAVs visitation of a single target from different positions, the models do not consider the path 
planning issues which are as necessary as the path planning especially for 
restricted environments. Multiple UAVs models have also been employed to visit many 
target \citep{tuyishimire2019clustered, 
ismail2018internet,tuyishimire2018optimal, ismail2018generating}. Here the 
focus was the efficient target visitation and the assignment issue has not 
been addressed. However when heterogeneous UAVs have to visit multiple sensors, an 
UAVs assignment model is required to complement path planning models.
 
This paper assumes a restricted and complex network, where sensors are not only connected in terms of their ability to forward data to each other, but also in terms of the possible paths UAVs may use to visit each sensor from any base station or any other sensor in a region of interest. We propose a persistent and  real-time path planning together with a task allocation model where, a team of  heterogeneous UAVs coming from various base stations are used to collect sensor readings from ground sensors and deliver the collected information to their closest base stations. The underlying data muling problem is i) mathematically formalised as a real-time constrained optimisation problem, ii) proven to be intractable and iii) solved using a novel heuristic solution, whose relative efficiency is proven through running experiments on both artificial and real sensor network, with various UAVs settings. 

This paper is an extension of the work in \citep{tuyishimire2017cooperative}. An extension has been done by  detailing the work and the proposed model has been transformed to make the model real-time respondent. Furthermore, analysis results corresponding to added features have been provided.

The rest of this paper is organised as follows. The cooperative data muling model is presented in Section \ref{sec1} and its  algorithmic solution provided in the same section. Simulated results are provided and discussed in Section \ref{sec4} while the  conclusion is drawn in Section \ref{sec5}.

\section{The Cooperative Data Muling Model} \label{sec1}

\begin{figure}[h!]
 \centering
 \includegraphics[scale=0.55]{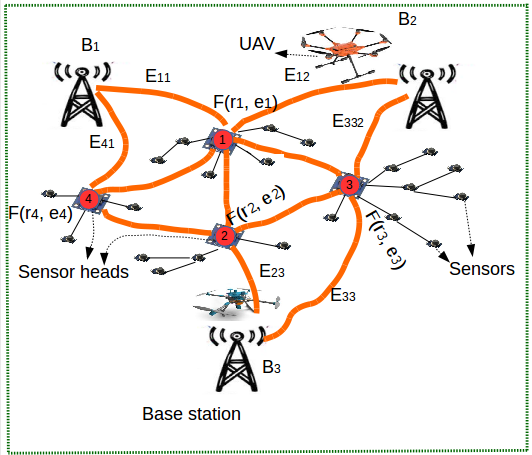} 
\caption{Cooperative Data Muling
 \label{iot-im3}}
 \end{figure} 
 
\FloatBarrier

 In this paper, we consider an ``Internet-of-Things in Motion'' model as shown in  Figure~\ref{iot-im3}. We assume that UAVs are assisted by special ground-based sensors which locally collect data from other sensors. That is, sensors are grouped into separated clusters, each with its own sink node (the cluster head), where  the information is to be collected from other sensor nodes (cluster members) and  relayed to UAVs which deliver the sensor readings to base stations. Note that only cluster heads  can communicate with UAVs, and the optimal clustering scheme is not covered in this paper. Furthermore, the inner cluster communication technology is not covered here (it has been discussed in \citep{tuyishimire2014internet,hope2019, mhope2019}).

The cooperative data muling model  considered in this paper is illustrated by Figure~\ref{iot-im3}. The figure reveals three base stations ($B_1$, $B_2$ and $B_3$) from where UAVs take off to collect data from sinks located in a region of interest and later comes back for data delivery. In this illustrative scenario, all possible collection paths which can be taken by each UAV from the base stations to access data collected by  cluster heads ($1$, $2$, $3$ and $4$) and deliver the collected data to the closest base stations, are represented in big and orange lines. Here, we assume that UAVs are assisted by special nodes (cluster heads/sinks) to collect information to reduce the loss due to UAVs capabilities (fuelling, timing, etc.). Furthermore, it is assumed that the UAVs paths topology is known (this means that all possible communication paths between all sensors are known) and it is represented by small black lines in the figure. 
The total energy required for data collection at a node/sensor, say $i$, is a function $F(r_i, e_i)$ of its revisit deadline $r_i$ (the maximum amount of time required to (re)visit the node) and the energy $e_i$ required 
to transfer data from node $i$ to an arriving UAV. The travel cost from a base station $B_j$ to a sink $i$ is translated in an energy metric denoted by $E_{ij}$, 
and the transportation cost between two sinks $i$ and $k$ is also translated into an energy metric denoted by $e_{ik}$.

\subsection{The Data Muling Problem}
In this subsection, the problem is modelled as a constrained optimisation problem. We start by defining/denoting all cost related terms and later, we  combine them to form a cost function.% The properties of a good system stated in Subsection \ref{gdsys}, are to be optimised.

\subsubsection{UAVs waiting time on sink nodes }
Let $t_i$ be the entire time spent by an UAV to arrive at  the sink $i$ since it took off from a base station and $r_i$ be the expected time for a UAV to arrive at the sink $i$. It is also referred to as {\it revisit time} at the collection point $i$. 

The sink visitation-based cost may be expressed in terms of penalties for both early and late visits on the sink nodes, the collection 
of information and a risk associated with the autonomy of the UAVs. These costs are described as follows.

\begin{itemize}
\item {\bf Early visit penalty:} An early visit penalty will be assigned to an  UAV if $t_i <r_i$ to express the case where the  visiting UAV arrive 
premature  data collection. In this case, the UAV will wait for a period of time  $w_i=r_i -t_i $ needed by the sink node to capture mature information from the field and transmit 
it to the waiting UAV. 

\item {\bf Late visit penalty:} A late visit penalty will be applied to the UAV if  $t_{i}>r_i$ to express the fact that the visiting UAV is late by a 
period of time $l_i=t_i -r_i$ wasted by the UAV to arrive late to a collection point where data was ready for collection.This penalty can also be expressed 
using a piece-wise function. 

\item {\bf Data collection cost:} A data collection cost will be applied to any UAV to consider the fact that  the UAV has to use 
the energy $e_i$ to collect information from the visited node $i$. Note that while the costs $w_i$ and $l_i$ depend on 	how the terrestrial and airborne sink networks have been traffic engineered, the data collection cost $e_i$ may depend on different engineering parameters and functions which may be bound to the communication interfaces of the equipment used by both the ground sink nodes and the UAVs and the protocol used for such communication. 
\end{itemize}

For each UAV, the total cost $F(r,e_i)$ of visiting  the sink $i$, without taking into account the travelling cost is expressed by  
\begin{equation}\label{nodeweight}
F(r_i,e_i) = \alpha w_i u(w_i)+\beta l_i u(l_i)+\gamma e_i,
\end{equation}
where $u(w_i)$ and  $u(l_i)$ are the values of a unit step function applied to $w_i$ and $w_l$,respectively. The coefficients $\alpha$, $\beta$, and  $\gamma$ are associated with the setting-based importance/weighting allocated to the early and late arrival penalties and the data transfer penalty respectively.

\subsubsection{The assumed network }\label{nt}

We consider a hybrid  sensor network ( a network with multiple types of links) that represented by a bi-directed graph $\mathcal{G = (S,N,L,B,P)}$, where $\mathcal{S}$ is the set of all sensor nodes, $\mathcal{N} \subset \mathcal{S}$ is the set of all sinks, $\mathcal{L}$ is the set of wireless communication links between the sensor nodes, $\mathcal{B}$  is the set of UAV base stations while $\mathcal{P}$  is the set links showing possible moves of UAVs. 
Here, a move  expresses one of the two following kinds of connection.
\begin{itemize}

\item \textbf{Base station-sink :}  These are bidirectional UAVs paths connecting sinks and base stations. The cost of moving from a base station $b$ to a sink $i$ is denoted $E_{bi}$ and its opposite is $E_{ib}$, with $E_{ib}=E_{bi}$.
\item \textbf{sink-sink :} These are the UAVs paths connecting sinks amongst themselves and the cost to move from one sink $i$ to $j$ is denoted by $e_{ij}$, with $e_{ij}=e_{ji}$.
\end{itemize}

\subsubsection{Initial condition }
\begin{itemize}

\item Each UAV is assumed to start its journey from a base station.
\item The waiting times at all sensor nodes. That is $l_i=w_j=0, \forall i,j \in \mathcal{N}$
\end{itemize}

Here, it is assumed that the maximum number of UAVs at each base station is equal to the degree/capacity of the base station.

\subsubsection{The data muling modelling} 

The data muling is performed in two steps

\begin{itemize}
\item \textbf{Data collection.} During data collection, an UAV  is to move from a Base station $a$ to collect data from  $k>0$ sinks labelled by a set of indices $p^*=[1,2,..., k]$ . In this case, we represent the path used for data collection by $p=[a,1,2,..., k]$ . The energy required for this step is expressed by 

\begin{equation}\label{sel}
C(p)= E_{a1}+\sum \limits_{i\in p^*}F(r_i,e_i)+\sum \limits_{i,j\in p^* \wedge j=i+1}e_{ij}.
\end{equation}
\item \textbf{Data delivery.}  During data delivery, a UAV may or many not pass by some already visited sink to deliver information to closest base station $b$. However, the end point of the collection path $p$ is the starting point of the delivery path. In this case, the corresponding energy is expressed as a function $E(p)$ of the data collection path.
\end{itemize}
Therefore, the total energy required for data collection and delivery is given by the equation
\begin{equation}\label{cost}
 C(p)+E(p)= E_{a1}+\sum \limits_{i\in p^*}F(r_i,e_i)+\sum \limits_{i,j\in p^* \wedge j=i+1}(e_{ij}) +E(p).
 \end{equation}
  
%Let $U=\{1,2,3,...m\}$ be a set of UAVs  distributed in a set $BS=\{1,2,3,..., n\}$
 %of the base station. The UAVs are to persistently and periodically visit the set of sink $S=\{1,2,3,...,k\}$.
 %Let $\mathcal{P}$ be a set of all paths which could be taken by UAVs. The set

The data muling problem consists of finding for each UAV an optimal path so that the total energy spent by all the UAVs to collect and deliver the sensor readings/data without colliding is minimized. Mathematically, we represent the set of UAVs by $U=\{1,2,3,...m\}$, where each UAV say $u$ departing from  base station $a_u$  will follow path $p_u$ to collect data at locations of interest and another path (maybe different from $p_u$)  to deliver the data to its closet base station. Let's consider $1_u$, the first sink to be visited by the UAV $u$. The data muling problem is formulated as follows.

\newpage
\begin{equation} 
 min\  Z= \mathlarger{\mathlarger{‎‎\sum \limits _{u=1}^m}}\bigg( E_{a_u1_u}+\sum \limits_{i\in p_u^*}F(r_i,e_i)+\sum _{\substack{i,j\in p_u^* \\ j=i+1}}(e_{ij})+E(p_u)\bigg),
 \qquad\qquad \label{eq:udm} 
\end{equation}

$$
\begin{array}{lll} 
subject 					& to 					& \mbox{}
\end{array}
$$

\begin{description} 
\item [{\bf (\ref{eq:udm}.1)}] $\forall v,w \in U,p^*_v \cap p^* _w=\emptyset = d(p^*_v)\cap d(p^*_w)$
\item [{\bf (\ref{eq:udm}.2)}] $\bigcup \limits_{u\in U}p_u^*=S$
\item [{\bf (\ref{eq:udm}.3)}] $e_i$, $e_{ij}$, $E(p)$, $r$, $E_{a_u1_u}$ $\geq 0$, $\forall$ $i$, $j$, $p$, $a_u$, $1_u$.
\end{description} 
Here, the first constraint states that any two collection or delivery paths have no sink in common. This guarantees collision avoidance for the UAVs. 
On the other hand, the second constraint expresses the fact that all sinks are to be visited.
  
\subsection{Real-time visitation}

We consider Equation \ref{nodeweight}. In the scenarios where the variables have  very strict conditions, instead of being part of the cost function, they need to be part of the problem constraints. Table \ref{t1} shows all possible models. Here a "1" shows the case where a corresponding variable is restricted (part of the constraints) and a "0" shows the other way.

  %\FloatBarrier
 \begin{table}[h!]
 \scriptsize
 \centering
 \begin{tabular}{|ccl|}
 \hline
 
 Waiting penalty ($ w_i$) & Late penalty ($l_i$)  & Explanation\\\hline
 0 & 0  & None of the two variables is bounded \\
 0 & 1  &  Only the late penalty is bounded \\
1 & 0  &  Only the waiting penalty is bounded \\
 1 & 1  &  Both variables are bounded\\\hline
 \end{tabular}
  \caption {Data collection scenarios.}
\label{t1}
 \end{table}
  % \FloatBarrier
The optimisation problem  changes its constraints so as to become as follows.

\newpage
 \begin{equation}  
 min\  Z= \mathlarger{\mathlarger{\sum}}\limits _{u=1}^m \bigg( E_{a_u1_u}+\sum \limits_{i\in p_u^*}F(r_i,e_i)+\sum _{\substack{i,j\in p_u^* \\ j=i+1}}(e_{ij})+E(p_u)\bigg),
\qquad\qquad \label{cost} 
\end{equation}

$$
\begin{array}{lll} 
subject 					& to 					& \mbox{}
\end{array}
$$

\begin{center}
 \begin{description} 
\item [{\bf (\ref{cost}.1)}] $\forall v,w \in U,p^*_v \cap p^* _w=\emptyset = d(p^*_v)\cap d(p^*_w)$
\item [{\bf (\ref{cost}.2)}] $\bigcup \limits_{u\in U}p_u^*=S$
\item [{\bf (\ref{cost}.3)}] $0\leq w_i\leq W_i$
\item [{\bf (\ref{cost}.4)}]  $0\leq l_i\leq L_i$
\item [{\bf (\ref{cost}.5)}] $e_i$, $e_{ij}$, $E(p)$, $r$, $E_{a_u1_u}$ $\geq 0$, $\forall$ $i$, $j$, $p$, $a_u$, $1_u$.
\end{description}
 \end{center} 
where $W_i$ and $L_i$ are the predetermined thresholds which may take any non negative value.
%%%%%%%%%%%%%%%%%%%%%%%%%%%%%%%%%%%%%%%%%%%%%%5

\subsection{Related problems and solutions}\label{rel}
The  data muling problem considered in this paper is closely related to the 
file recovery problem in \citep{pal2009evolution} 
(NP-hard problem) solved  by curving techniques, including those using the Parallel Unique Path (PUP) algorithm. This problem 
considers a case of many fragmented files which need to be reassembled, starting from their headers, which are assumed to be known 
initially. The PUP algorithm  is a variation of Dijkstra's routing 
algorithm\citep{dijkstra1959note} where starting from the headers, clusters are successively added based on their  best matches. 
This is done with the aim of building paths from headers having a cluster added to an existing path if and only if the link to it 
has the least weight. On the other hand the Vehicle Routing Problem (VRP)\citep{laporte1992vehicle,toth2001overview} consists of finding the optimal road from a depot, to be taken for delivering resources to customers and return to the depot. Exact and heuristic algorithms for its solution have been surveyed 
in\citep{laporte1992vehicle}. In the survey, all stated algorithms assume a single distance matrix (the  cost matrix)  and hence 
could fail to be a good fit for our persistent visitation scenario since in our case the weighting of nodes matters and it is not a 
fixed value. Furthermore, for the VRP, vehicles end their trips at the depots where they started from. This would limit the number of topologies where the data muling problem is solvable and also could impose a data muling scheme which is not necessarily optimal. Note that in our case, we are interested in the case where the late and stale visitation are taken care of, and this depends on the dynamic position of UAVs (see the Equation \ref{nodeweight}). Furthermore, UAVs deliver the collected information to  optimal base stations (which are not necessarily their starting points).

\subsection{The data muling problem intractability}
 
To prove its intractability, we provide a polynomial reduction of one-depot VRP  (which is known to be an NP-hard problem), into a special case of the data muling problem: the case where each sink's weight is zero. The transformation consists of a two-step process which transforms the graph $\mathcal{G}$  as follows.
\begin{itemize}
\item [\textbf{a.}] Group all Base stations in $\mathcal S$  in one cluster/group and consider this cluster/group as a special node for the graph, this gives the VRP's topology $\mathcal{G'= (S,N,L,B',P)}$, where $\#\mathcal{B'}=1$. 
\item [\textbf{b.}] For every link of $\mathcal{G'}$, make the link weight in the new graph (found in \textbf{a.}) the inverse of the weight in the Graph ${\mathcal G'}$.
\end{itemize} 
 This will reduce the VRP's into the data muling problem's solution.
 
Clearly, the time complexity of the transformation process is polynomial since Step $a$ has a complexity $\mathcal{O}(\#\mathcal S)$ and  Step $b$ has complexity $\mathcal{O}(\#R)$. The time complexity for the whole graph transformation/reduction process is therefore 
$\mathcal{O}(\#R) +\mathcal{O}(\#\mathcal S)$, which is polynomial. This shows that the problem of interest in this paper is  NP-hard and hence a heuristic solution is important.

\subsection{The Data Muling Algorithm}

 In this work, we adapt  Dijkstra's algorithm, in the same way it is done in \citep{pal2009evolution}, to solve the data muling problem. While many rounds 
 are considered by our algorithm, we consider only the case where each node is visited only once per round. It is assumed that each UAV is capable of 
 collecting and delivering data to a base station where it can be recharged, before going for another data collection round.

\begin{algorithm}[h!]

 \small
\textbf{Assume} a network G(N,L,B,P) as specified in Section \ref{nt}\;
%Let ${\mathcal Q}$ be the dictionary of UAVs and  positions of sink being visited by them\;
%Let $Paths \in \mathbb{P}[\mathcal{Q}]$ be the dictionary consisting of each UAV's paths to its position (a base station)\;
%Let ${\mathcal S}$ be the set of all the base stations\;
$Choices \leftarrow$  all sink to be visited \;
Initialise the path to the initial hosting base stations\;
$done \leftarrow choices \cup B$\;
\While {$choices \neq \emptyset$}{
	%Assign=\{\} \tcp{drones and their possible assignments}
	\For{$ u \in U$}{
		Select the next destination of least cost, using Equation \ref{sel} \;% \tcp{neighbour to be visited in a shortest time}\\
	%Assign[u] $\leftarrow$ n\;
}
	$Assign=\emptyset$\;
	\For {$u \in U$}{
		Let $c$ be the choice of $u$\;
		\For {$v \in U \setminus Assign$}{
			\If {$u$ and  $v$ selected the same choice $c$}{
				Include $c$ in the path of a UAV of least cost (\ref{sel})\;
				Include the UAV in $Assign$\;
				Include the $c$ in $done$\;
			 	$Choices \leftarrow Choices \setminus \{c\}$\;
			 	Break\;
			 	}

				}
			 \If {$c\in Choices$}
			{
				Include $c$ in the path of  $u$\;
				Include the $c$ in $done$\;
				$Choices \leftarrow Choices \setminus \{c\}$\;
				Include $u$ in $Assign$\;
				Break\;
		}

}
	\uIf { $Choices = \emptyset$ or $\#done=\#U$}{
	
	$Choices \leftarrow B$\;
	$done \leftarrow \emptyset$\;
	$B \leftarrow \emptyset$\;
	\If  {$\#done \neq \#U$}{
		$Cpaths \leftarrow$ all UAVs' paths\;}
	}	
	
}
Dpaths $\leftarrow$ all current UAVs' paths \;
Return Cpaths and Dpaths
 \caption {Cooperative algorithm \label{algo}}
\end{algorithm}

Algorithm \ref{algo} has two major steps: the first step consists of using Equation \ref{sel} to select the best node to visit for  every UAV (Step 6-7); the second step consists of adjusting the UAV's paths by choosing the cheapest UAV  for every best node (Step 8-23).
Once the visitation is done, the collection paths are captured in $Cpaths$ and the two steps are repeated to select the nearest base station for every UAV data delivery following the delivery paths recorded by $Dpaths$.

\FloatBarrier
%\subsubsection{Algorithm Analysis}

\begin{pro}[\textbf{Polynomial termination}]
Algorithm \ref{algo} terminates in polynomial time, when all sink nodes have been visited.
\end{pro}

\begin{pr}
Note that  each time a sink is included in a path of one of the UAVs, it gets excluded from the list $choice$ (Lines 16 and 21). Since all  UAVs paths consist of a connected graph, whenever $choice \neq \emptyset$, there is at least one UAV which makes a new selection of a next sink to visit on Line 6. So all sink are visited.

 Once $choice = \emptyset$,  the next destinations become the base stations  and the set $done=\emptyset$ (Lines 25 and 26 consecutively). The next step is to make $\#done=\#U$ true by assigning to every UAV a base station. In this case the statement at Line 24 is true and the set $choice=B$ which had been updated to $\emptyset$. This makes Algorithm \ref{algo} stop.
 
 On the other hand, the time complexity of the algorithm is clearly $\mathcal{O}((\#U)^2)$, which is a polynomial. Hence the result follows.
\end{pr}

\begin{rem}[\textbf{Persistent visitation model}]
Since each UAV's computed path is ended by a base station and the UAV paths form a static network, Algorithm \ref{algo} can be used repetitively to make a persistent visitation.
\end{rem}

\FloatBarrier
%
%
%\subsection{Algorithm}\label{algo}
%In this work we adopt  Dijkstra's algorithm \citep{dijkstra1959note}, in the same way it is done in \citep{pal2009evolution}, to solve the problem described in Subsection \ref{rel}. We present the algorithm which runs in rounds. We consider the case where each node is visited only once per round.
%\begin{enumerate}
%
%%\item For each UAV, order the neighbouring sensors in terms of cheapest to visit. Here, the cost to visit one node is the travelling cost plus the weight of the node. 
%%\item For all UAVs, assign the cheapest move and remove it from the further choices (a choice  an unvisited  sensor).
%%\item  Repeat Step 2 until no more choice is available (all sensor nodes have been visited).
%%\item  For each UAV position assign the best base station. That is the base station to which, the travel cost is least.
%
%
%\item For each UAV, select the best (cheapest) sensors to visit. Here,
%the cost to visit one node is the travelling cost plus
%the weight of the node.
%
%\item For all selections, choose the cheapest one, make it
%the current assignment and remove the chosen sensor
%from the selectable ones.
%
%\item Repeat Step 2 until no more choice is available (all
%sensor nodes have been selected).
%
%\item  For each UAV position assign the best base station. That is the base station to which, the travel
%cost/distance is least.
%
%\end{enumerate}

\subsection{Illustration of the algorithm}

\begin{figure}[h!]
 \begin{subfigure}[b]{0.48\textwidth}
 \centering
 \includegraphics[scale=0.3]{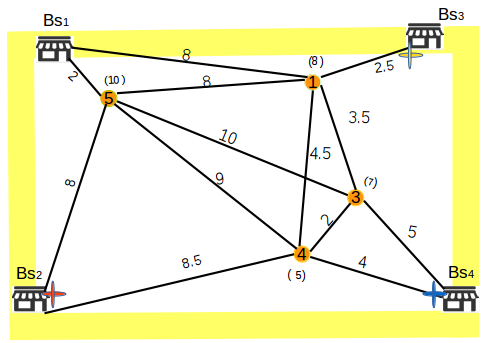} 
  \caption{Initial step
 \label{model1}}
\end{subfigure}
 \begin{subfigure}[b]{0.48\textwidth}
 \centering
 \includegraphics[scale=0.3]{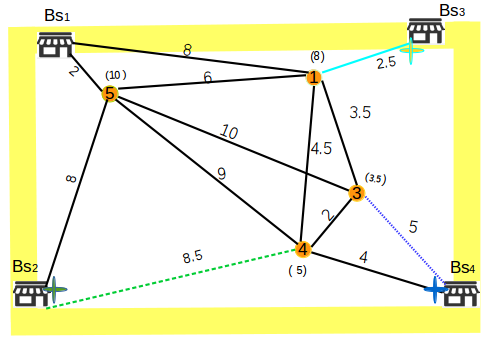} 
  \caption{The best neighbour choice.
 \label{model2}}
\end{subfigure}

 \begin{subfigure}[b]{0.48\textwidth}
  \centering
 \includegraphics[scale=0.3]{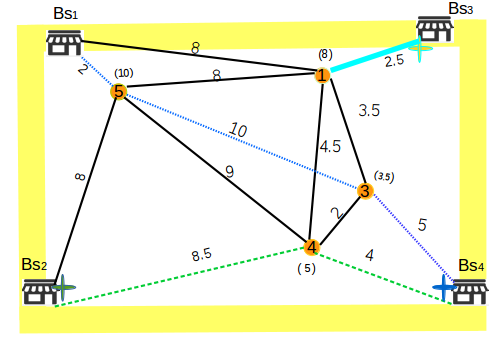} 
 \caption{Delivery.
 \label{model3}}
\end{subfigure}

 \caption{Illustrative example.}
 \label{i_example}
 \end{figure}

We illustrate the algorithm in Subsection \ref{algo} using an example in Figure \ref{i_example}. We run  one round of the algorithm step by step. In this example we consider a case were each sensor node weighting is constantly zero. That is,  $\alpha=\beta =0$ (see the constants in  Equation \ref{nodeweight}).

 Figure \ref{model1} shows the initial conditions of the considered system. The system has  four sensors, four base stations and three UAVs positioned at all of them except Base station $Bs_1$. The links and sensors are weighted as discussed in Section \ref{sec1}.
 Let $u_2$, $u_3$, and $u_4$ be the name of UAVs staying at  Base Stations $Bs_2$, $Bs_3$, and $Bs_4$, respectively.

Figure \ref{model2} reveals how the UAVs make choices of their first moves. The best choice is the one corresponding to the cheapest move, which is evaluated using the weight of the road to be used together with the delay at the sensor to be visited. This is why UAVs $u_2$, $u_3$ and $u_4$ move to Sensors $4$, $1$ and $3$, respectively. At this stage, only node $5$ is the only node not yet visited.

Figure \ref{model3} shows the next moves up to the end of the algorithm. It shows that UAV $u_4$ moves to Sensor $5$, because it is the one corresponding to the cheapest move. On the other hand, all other UAVs do not have any other choice of sensor to visit. They then need to visit their closest base station. Once UAV $u_4$ arrives at Node $5$, it visits the closest base station which is $Bs_1$.

\section{ Experimental results}\label{sec4}

\begin{figure}[h!]
 \begin{subfigure}[b]{0.48\textwidth}
 \centering
 \includegraphics[scale=0.31]{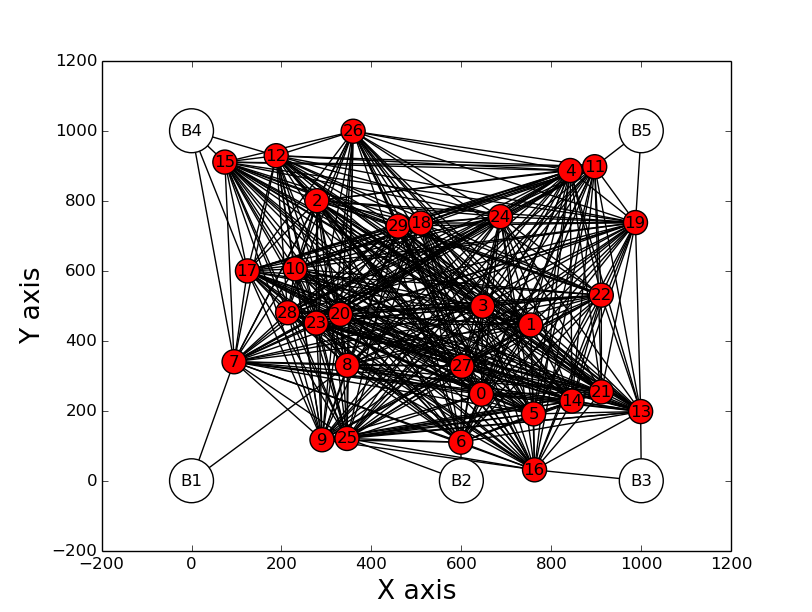} 
  \caption{Artificial network.
 \label{netw1}}
\end{subfigure}
 \begin{subfigure}[b]{0.48\textwidth}
\centering
 \includegraphics[scale=0.4]{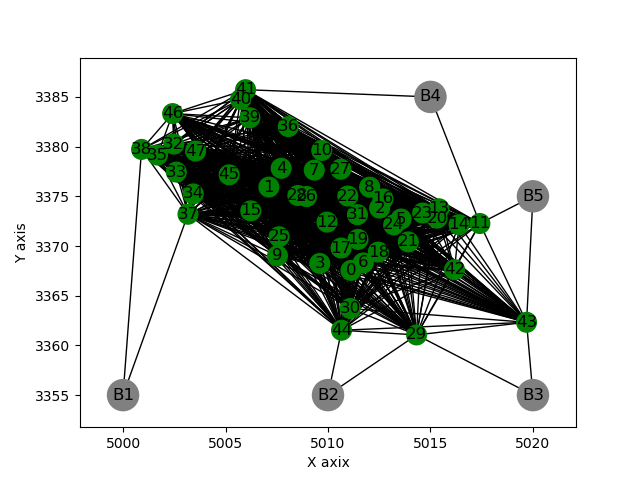} 
 \caption{Real Cape Town  network. 
\label{netw2}}
\end{subfigure}
 \caption{Considered networks.}
 \end{figure}

 Python was used to run  Algorithm \ref{algo} on two more complex networks (Figures \ref{netw1} and \ref{netw2}). The performance of the algorithm is studied and the behaviours of considered parameters are investigated. 
 The first  network (\ref{netw1}) consists of five base stations: $B_1$, $B_2$, $B_3$, $B_4$ and  $B_5$, as shown by bigger nodes in Figure \ref{netw1}. In the figure, smaller  nodes represent the sink to be visited and the links in the network show the possible paths, the UAVs may take to visit the targets. Note here that the sink's network (network without base stations) is a complete graph (each UAV is able to move from one sink to any other one in the network, but not to any base station) where nodes are randomly deployed on a $1 km^2$ area. 
 
 On the other hand, we consider a real network (Figure \ref{netw2}) consisting of the Cape Town complete network whose nodes are police stations and their Cartesian coordinates have been extracted from GPS positions. The names corresponding to each node label are described in Appendix  (see Figures \ref{net00} and \ref{net10}).
  Note here that in these experiments the considered UAV are drones.

As shown in Figure \ref{netw1}, the considered network consists of nodes randomly placed in an area of size $1km^2$ and sinks are labelled in terms of the energy required to collect information from them. The coordinates of nodes in both networks are in metres and could be seen or approximated using Figures \ref{netw1} and \ref{netw2}. Positions (of sinks or base stations) consist of triples but for simplicity of plotting/viewing them, they have been  projected  on X-Y coordinates and hence, they are presented in the  $2D$ Cartesian coordinate system.

 %\FloatBarrier
\subsection{Impact of speed distribution on path planning}

\begin{figure}[ht]
 \begin{subfigure}[h]{0.48\textwidth}
  \centering
 \includegraphics[scale=0.3]{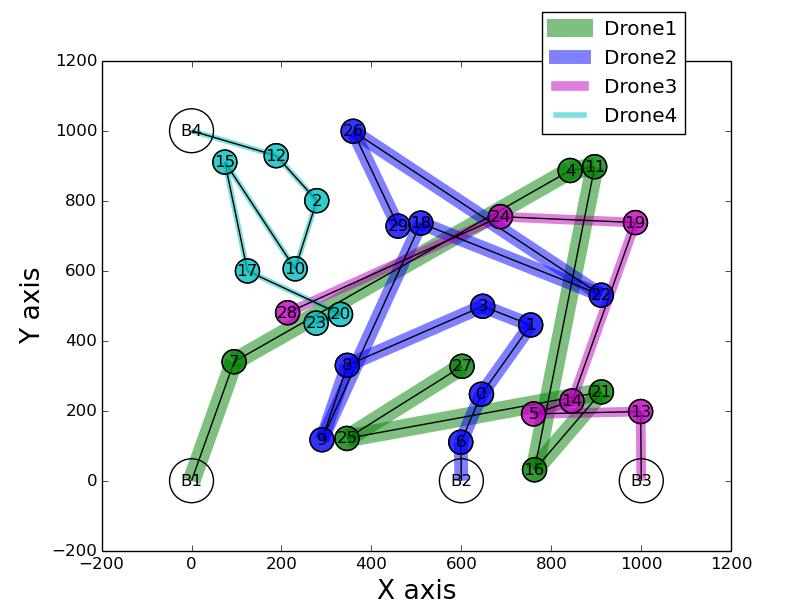} 
  \caption{Path generation on a random network when the UAVs have the same speed.
 \label{r1}}
\end{subfigure}
 \begin{subfigure}[h]{0.48\textwidth}
\centering
 \includegraphics[scale=0.4]{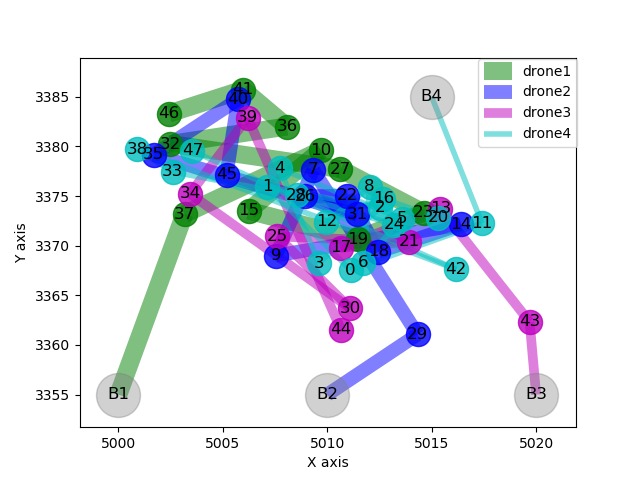} 
 \caption{Path generation on a real network when the UAVs have the same speed. 
\label{rn}}
\end{subfigure}
\caption{Path generation.}
\end{figure}
%\FloatBarrier
 
In two steps, we present the paths taken by UAVs using Algorithm \ref{algo}.

\begin{itemize}
\item[\textbf{Step1.}] \textbf{Data collection:} it consists of visiting all sinks using the first three steps of Algorithm \ref{algo}. The  corresponding path for each UAV is shown in Figures \ref{r1}, \ref{rn}.% and \ref{r3}.

\item[\textbf{Step2.}] \textbf{Data delivery:} it consists of visiting  base stations using the last step of the same algorithm. The results are shown in Tables \ref{t22now}, \ref{t2now} and \ref{t3}, where the speed distribution of drones is also presented.
\end{itemize}

The assumed cost function parameters are set to $\alpha=\beta =0.5$ and $\gamma =1$.

 Figures \ref{r1} and \ref {rn} reveal that UAVs do not visit the same number of sinks in the case of both considered networks. For example in Figure \ref{r1}, Drone1, Drone2, Drone3 and  Drone4 visit 7, 10, 6 and 7 sinks, respectively.
 % \FloatBarrier
 \begin{table}[h]
   \small
 \centering
 \scalebox{0.7}{
\begin{subtable}[t]{0.77\textwidth}
       \begin{tabular}{|p{1.4cm}p{1.8cm}p{1.5cm}p{1.8cm}|}
 \hline
 UAV name & Speed(m/min) & Source & Returning path\\\hline
Drone1  & 500 & B1 & [27, 6, B2]\\
Drone2  & 500 & B2 & [29, 12, B4]\\
Drone3  & 500 & B3 & [28, 7, B1]\\
Drone4  & 500 & B4 & [23, 8, B1]\\\hline
 \end{tabular}
  \caption{Delivery in the random network.}
 \label{t11}
 \end{subtable}
\begin{subtable}[t]{0.77\textwidth}
       \begin{tabular}{|p{1.4cm}p{1.8cm}p{1.5cm}p{1.8cm}|}
 \hline
 UAV name & Speed(m/min) & Source & Returning path\\\hline
Drone1  & 500 & B1 & [46,41, B4]\\
Drone2  & 500 & B2 & [45, 11, B5]\\
Drone3  & 500 & B3 & [44, B2]\\
Drone4  & 500 & B4 & [47, 12, B4]\\\hline
 \end{tabular}
   \caption{Delivery in the real network.}
 \label{t12}
 \end{subtable}}

 \caption{Delivery when all speeds are the same.}
  \label{t22now}
  \end{table}
%  {Data delivery when all drones have the same speed.}
%\label{t1}
% \end{table}
%\FloatBarrier

 Table \ref{t22now} shows that when the speeds are the same for all UAVs, the delivery requires some UAVs to pass by some of the visited nodes to arrive at the base stations. This is because each sinks does not need to be connected at a base station. Table \ref{t11} shows that in the case of random network,  Drone3 and Drone4 deliver the collected data at the same base station ($B_1$), and for the real network, Table \ref{t12} reveals that Drone1 and Drone4 deliver the data to B4.

%-------------------------------------------------------------------------------------------------------

% \FloatBarrier
\begin{figure}[h!]
\centering
\begin{subfigure}{0.48\textwidth}
  \centering
 \includegraphics[scale=0.3]{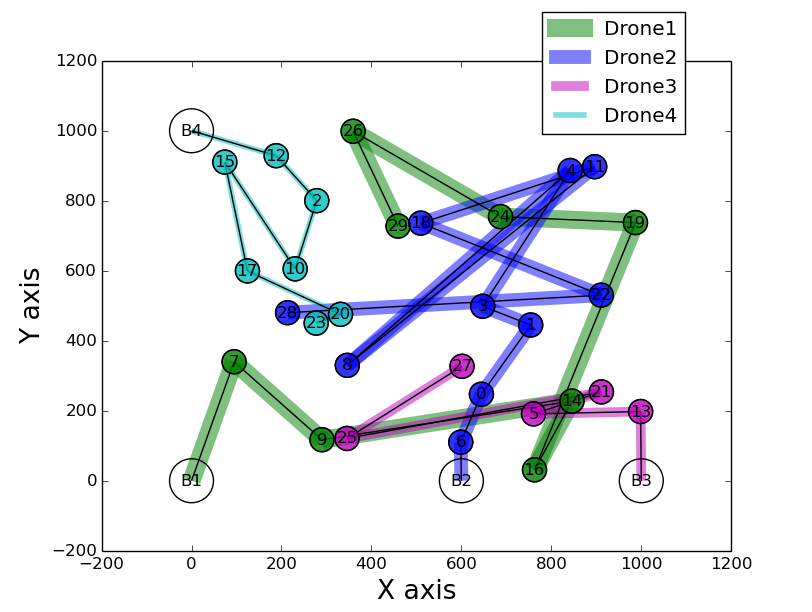} 
  \caption{Paths generation when UAVs have different speeds.
 \label{r2}}
\end{subfigure}
\begin{subfigure}{0.48\textwidth}
\centering
 \includegraphics[scale=0.4]{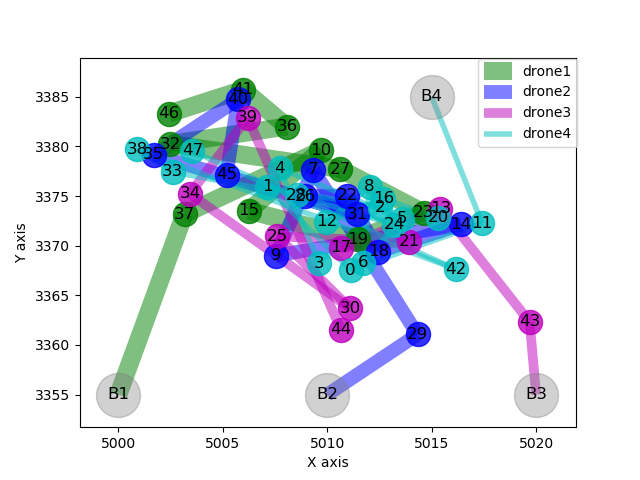} 
 \caption{Real network. 
\label{rn2}}
\end{subfigure}
 \caption{Paths generation when the UAVs have the different speeds.}
 \label{r2now}
\end{figure}
%\FloatBarrier 

Figure \ref{r2now} shows that when speeds are different, some UAVs may not change their paths but but other not. For example Drone4 does not change its path in both Figures \ref{r2} and \ref{rn2}, whereas all other drones do. On the other hand Figure \ref{rn2} shows that for the real network, all drones keep their paths.

  %\FloatBarrier
 \begin{table}[h!]
 \small

 \centering
  \scalebox{0.7}{
\begin{subtable}[t]{0.77\textwidth}
       \begin{tabular}{|p{1.4cm}p{1.8cm}p{1.5cm}p{1.8cm}|}
 \hline
 UAV name & Speed(m/min) & Source & Returning path\\\hline
Drone1  & 800 & B1 & [29, 12,  B4]\\
Drone2  & 700 & B2 & [28, 7, B1] \\
Drone3  & 600 & B3 & [27, 6, B2]\\
Drone4  & 500 & B4 & [23, 8, B1]\\\hline
 \end{tabular}
  \caption{Random network.}
\label{t21} 
 \end{subtable}%\hspace{2.5 cm}
\begin{subtable}[t]{0.77\textwidth}
       \begin{tabular}{|p{1.4cm}p{1.8cm}p{1.5cm}p{1.8cm}|}
 \hline
 UAV name & Speed(m/min) & Source & Returning path\\\hline
Drone1  & 800 & B1 & [46,41, B4]\\
Drone2  & 700 & B2 & [45, 11, B5]\\
Drone3  & 600 & B3 & [44, B2]\\
Drone4  & 500 & B4 & [47, 12, B4]\\\hline
 \end{tabular}
  \caption{Real network.}
 \label{t22}
 \end{subtable}}
  \caption{Data delivery when all drones have the different  speed.}
\label{t2now}
 \end{table}
%\FloatBarrier

 Table  \ref{t2now} shows that when the speeds are different, the delivery also requires some UAVs to pass by some of the already visited nodes, in order to arrive at a closest base stations. Table \ref{t21}  shows that Drone2 and Drone4 deliver the collected data at the same base station ($B_1$), and Table \ref{t22}  shows that Drone1 and Drone4 deliver the collected data at the same base station ($B_4$).

 %\FloatBarrier

%------------------------------------------------------------------------------------

% \FloatBarrier
  \begin{figure}[ht!]
 \centering

 \end{figure}

 %\FloatBarrier
\begin{figure}[h!]
\centering
\begin{subfigure}{0.48\textwidth}
  \centering
  \includegraphics[scale=0.3]{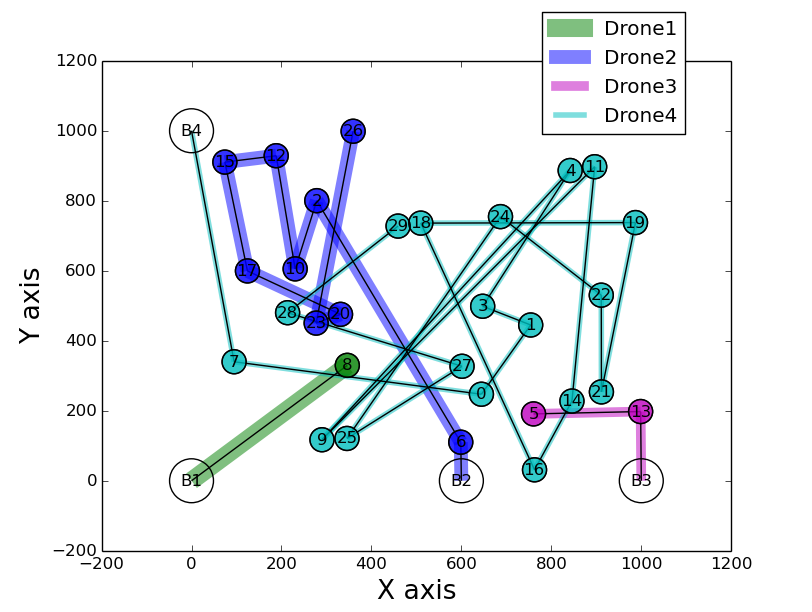}  
  \caption{Random network.
 \label{r33}}
\end{subfigure}
\begin{subfigure}{0.48\textwidth}
\centering
 \includegraphics[scale=0.4]{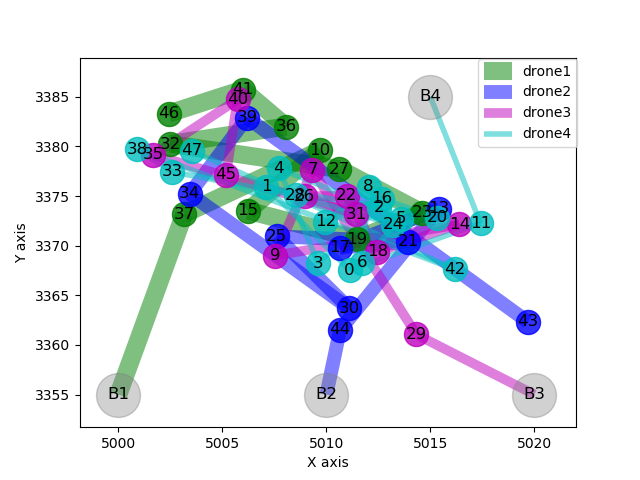} 
 \caption{Real network. 
\label{rn3}}
\end{subfigure}
  \caption{Paths generation when speeds distribution is changed.
 \label{r3}}
\end{figure}
%\FloatBarrier 

It is shown in Figure \ref{r3}, that the distribution of the UAVs speeds, have an impact on paths generation. For example Figure \ref{r33} shows that  Drone1, Drone2, Drone3 and  Drone4 visit 1, 9, 2 and 18 sink, respectively. This shows a big difference due to the fact that the choice of target depends on the current and not the previous visitation costs, together with the change of speed distribution in base stations.

   %\FloatBarrier
 \begin{table}[h!]
\small
 
 \centering
  \scalebox{0.7}{
\begin{subtable}[t]{0.77\textwidth}
       \begin{tabular}{|p{1.4cm}p{1.8cm}p{1.5cm}p{1.8cm}|}
 \hline
 UAV name& speed (m/min)& Source & Returning path\\\hline
Drone1  & 500 & B1 & [8, 6, B2]\\
Drone2  & 600 & B2 & [26, 12, B4]\\
Drone3  & 700 & B3 & [5, 16, B3]\\
Drone4  & 800 & B4 & [29, 15, B4]\\\hline
 \end{tabular}
     \caption{Random network.}
   \label{t31}
    \end{subtable}
\begin{subtable}[t]{0.77\textwidth}
       \begin{tabular}{|p{1.4cm}p{1.8cm}p{1.5cm}p{1.8cm}|}
 \hline
 UAV name & Speed(m/min) & Source & Returning path\\\hline
Drone1  & 500 & B1 & [46,41, B4]\\
Drone2  & 600 & B2 & [43, B3]\\
Drone3  & 700 & B3 & [45, 11, B5]\\
Drone4  & 800 & B4 & [47, 12, B4]\\\hline
 \end{tabular}
     \caption{Real network.}
   \label{t32}
 \end{subtable}}
    \caption{Data delivery  when speed distribution changes.}
   \label{t3}
 \end{table}
%\FloatBarrier

   Table  \ref{t3} shows that when the speeds are differently distributed, the paths are changed and thus the delivery paths also change. Table \ref{t31}  shows that for the random network, Drone2 and Drone4 deliver the collected data at the same base station ($B_4$); and for the real network, Table \ref{t32} shows that  Drone1 and Drone4 deliver the collected data at the same base station ($B_4$)

Since the path generation for both network essentially behave the same, we consider (the artificial) random network for the next analysis.

\subsection{Impact of the speed distribution on the cost (total energy)}

\begin{figure}[h!]
  \begin{subfigure}{0.49\textwidth}
 \centering
 \includegraphics[scale=0.32]{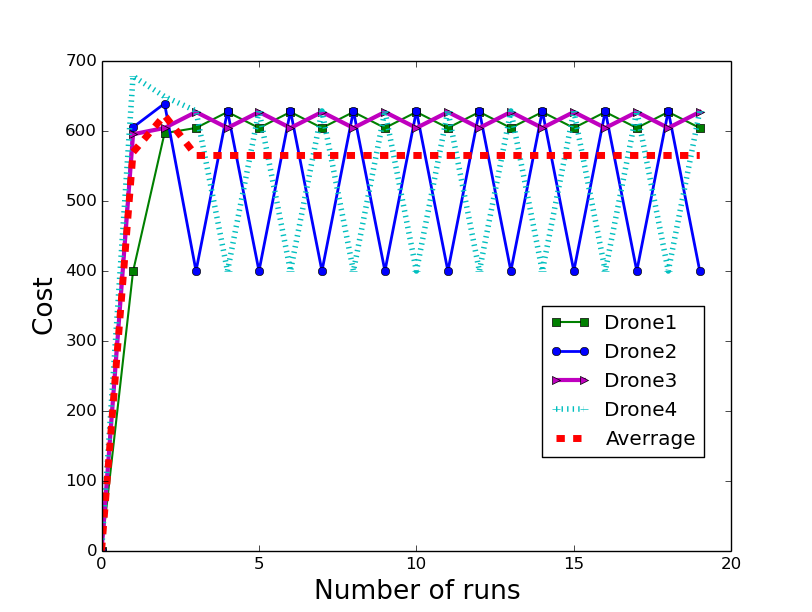} 
  \caption{Cost at same speed.
 \label{s1}}
 \end{subfigure}
\begin{subfigure}{0.49\textwidth}
 \centering
 \includegraphics[scale=0.35]{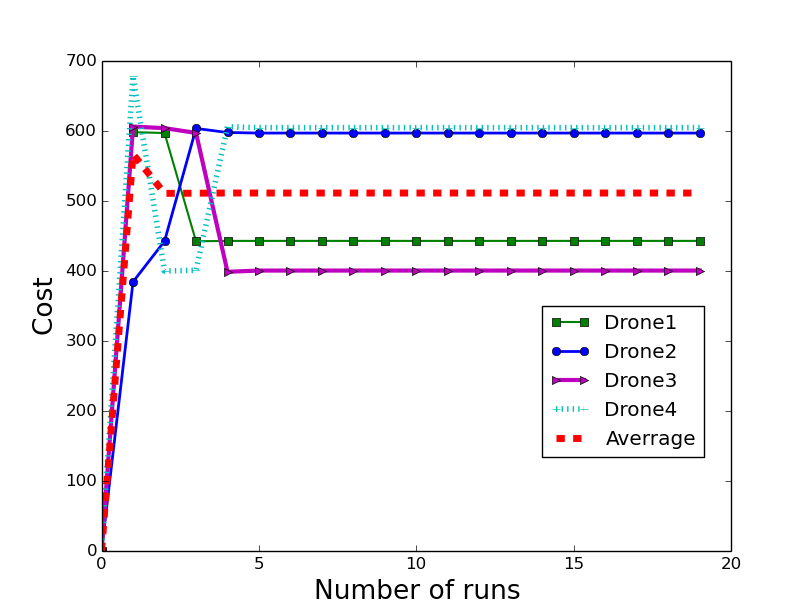} 
  \caption{Cost at different UAVs' speeds.
 \label{s2}}
 \end{subfigure}
   \begin{subfigure}{0.49\textwidth}
 \centering
 \includegraphics[scale=0.4]{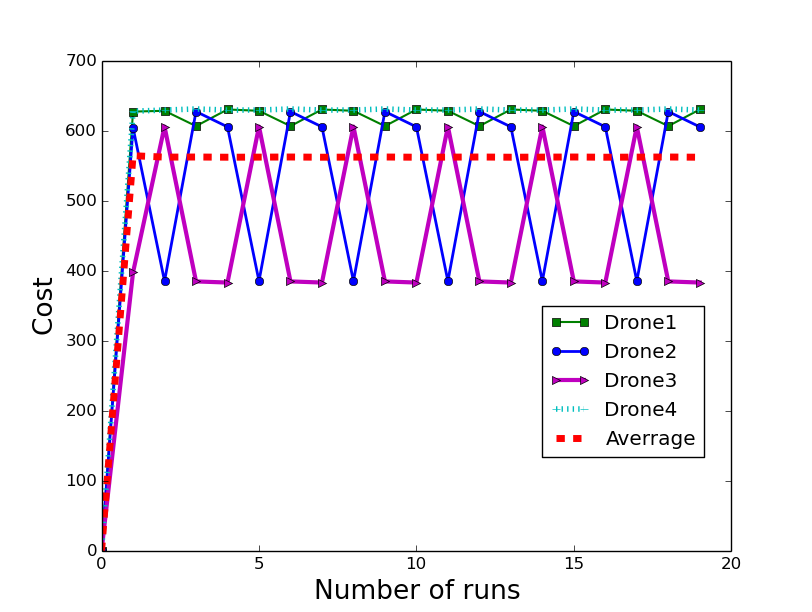} 
  \caption{Cost related to a different speed distribution.
 \label{s3}}
 \end{subfigure}
  \caption{Impact of speed distribution.}
 \end{figure}

We now study the impact of speed distribution on the cost considering many runs of the algorithm. We perform 20 runs  of Algorithm \ref{algo} in three cases of speed distribution, as shown by the second columns, in Tables \ref{t22now}, \ref{t2now} and \ref{t3}.
%\subsubsection{Same speed case} 
 Figure \ref{s1} shows  a case where each UAV' s speed equals 500m/min, and on the other side Figure \ref{s2} corresponds to the case where UAVs have different speeds as shown in Table \ref{t22now}.
%\FloatBarrier

  \FloatBarrier

 Figure \ref{s1} reveals that the cost for each drone lies in one of few fixed values. Drone4 takes four values and all others take three and succeed each other to take the minimum and maximum values. The average cost is constantly close to 560.
 
 %\subsubsection{Different speed case}

% \FloatBarrier

Figure \ref{s2} shows that after the first four  runs, all UAVs correspond to constant costs where the average cost is constantly close to 500. 
 
 %\subsubsection{Effect of speed distribution change}
% \FloatBarrier

 Figure \ref{s3} shows that the change in speed distribution may change the average cost and also the trends of the cost function. In this case, the amended speed distribution corresponds to the one in Table \ref{t3}, and  shows that for each UAV, the cost changes periodically and can only take one of a few fixed values. This is why the average cost also takes one of the fixed values on a periodic basis.
% \FloatBarrier 
 \subsection{Effect of parameters variation}

In this subsection, we study the  effect of four parameters on the cost variation as the number of runs varies. The four considered  parameters are described as follows.
 
 \begin{itemize}
 \item [$\bullet$] \textbf{Speed value}: keeping the speed the same for all UAVs, we aim to study the impact of its increase on  the coverage cost.
 \item [$\bullet$] \textbf{Overdue time ($\alpha$).} we study the impact of overdue time on the cost. Great attention is placed on this time by incrementing the corresponding coefficient (penalty) by 0.05, for each new run.
 \item [$\bullet$]\textbf{Delay ($\beta$).} While all the other parameters are constant, We vary the parameter corresponding to the delay penalty and study how it changes the value of the coverage cost.
 \item [$\bullet$] \textbf{ Data collection rate ($\gamma$).} Data collection rate is incremented by 0.05 for its value $\gamma = 1$, of 20 runs of the algorithm.
 \end{itemize}

  %\FloatBarrier
  \begin{figure}[ht!]
  \begin{subfigure}{0.49\textwidth}
 \centering
 \includegraphics[scale=0.32]{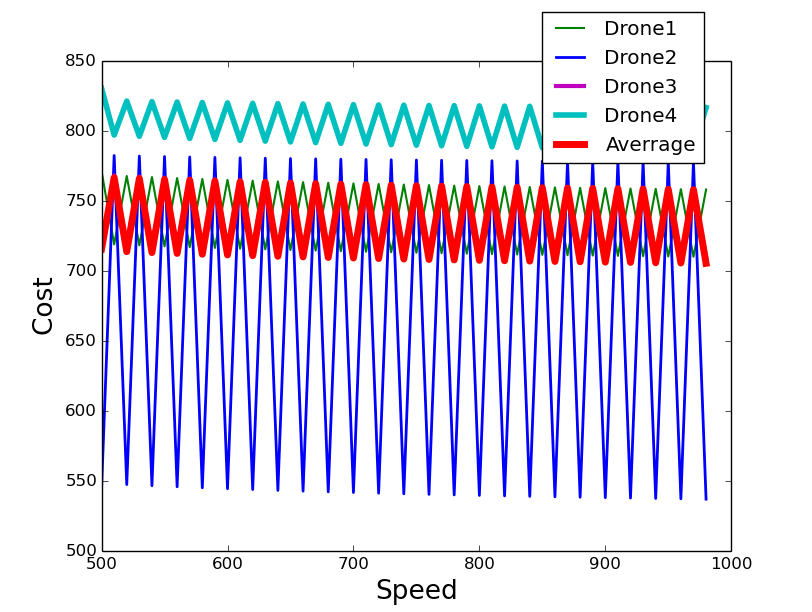} 
  \caption{Variation of the cost with respect to the speed.
 \label{C1}}
 \end{subfigure}
\begin{subfigure}{0.49\textwidth}
 \centering
 \includegraphics[scale=0.34]{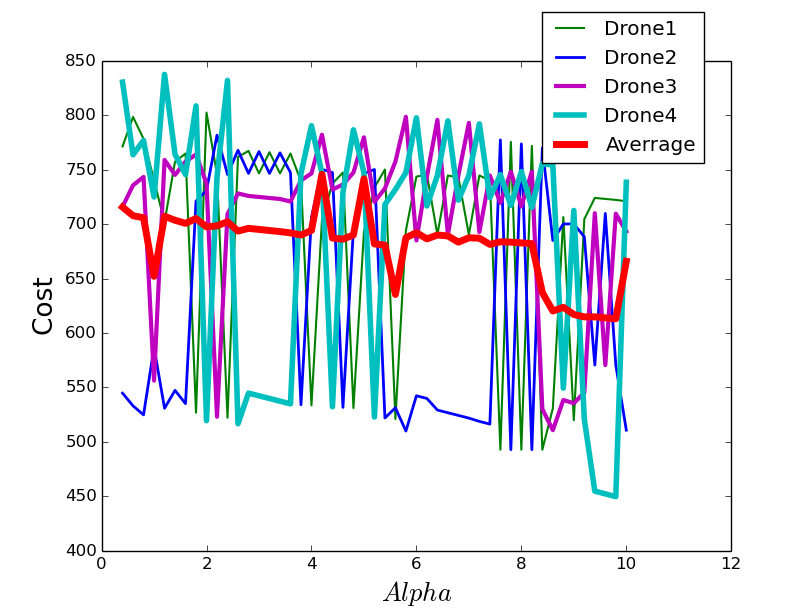} 
  \caption{Variation of the cost with respect to the overdue time ($\alpha$)
 \label{C2}}
  \end{subfigure}
  \caption{Impact of Speed and overdue time on the total cost.}
  \end{figure}
  
  \FloatBarrier
  
 Figure \ref{C1} shows a case where the speed has been incremented 20 times for all UAVs. It shows that from the first run, each UAV periodically changes its cost between two cost values. Drone4 corresponds to the highest cost while notably Drone3 corresponds  exactly to the average cost.

   Figure \ref{C2} shows a case where the overdue ( $\alpha$) is incremented  by 0.5, at each of 20 consecutive runs. The values of cost corresponding to each UAV is stochastic and the  average is stochastically decreasing.
% \FloatBarrier
  \begin{figure}[ht!]
   \centering
  \begin{subfigure}{0.49\textwidth}
 \includegraphics[scale=0.32]{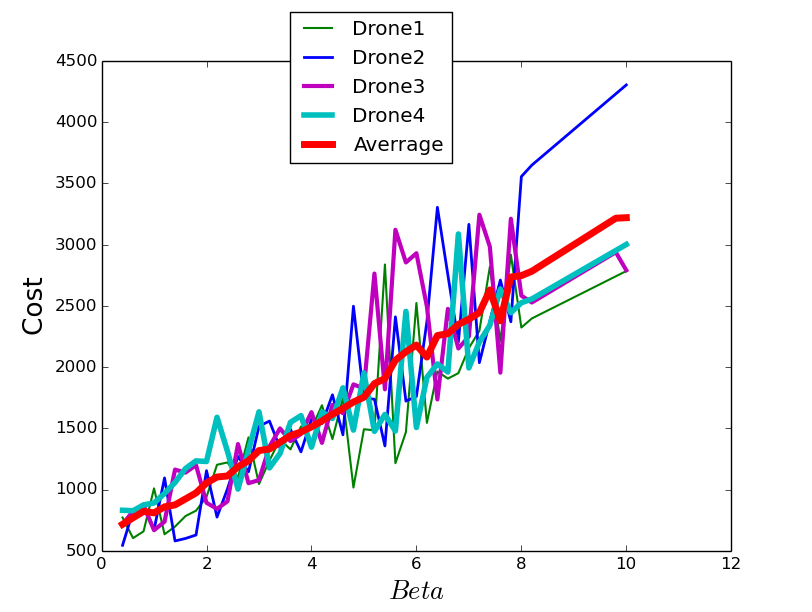} 
  \caption{Variation of the cost with respect to the  delay ($\beta$)
 \label{c3}}
 \end{subfigure}
 \begin{subfigure}{0.49\textwidth}
 \includegraphics[scale=0.35]{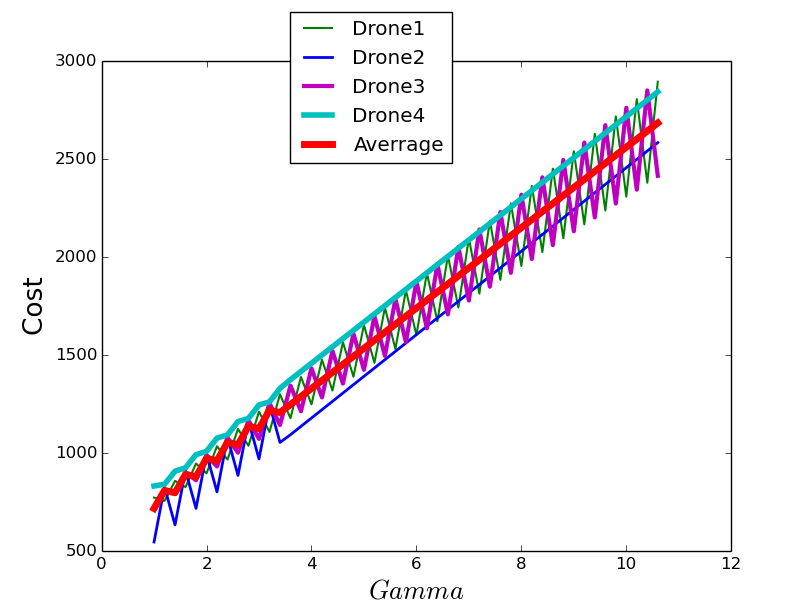} 
  \caption{Variation of the cost with respect to the data collection rate.
 \label{cc4}}
  \end{subfigure}
   \caption{Impact of the delay and data collection rate on the total cost.}
 \end{figure}
    %\FloatBarrier

We consider a case where the latency is the factor that changes in Figure \ref{c3}.
 The figure shows stochastic values as well but however the average cost is mostly increasing.

 Figure \ref{cc4} shows the impact of $\gamma$ on the variation of the cost over 20 runs. It shows that Drone4 mostly corresponds to the highest cost, and once $\gamma>3.8,$ Drone4 and Drone2 are constantly increasing their corresponding costs, whereas the others are periodically increasing and decreasing.
 %\FloatBarrier

  %\FloatBarrier
  \subsection{Prioritisation analysis}
  
  In this subsection, we discuss the effect of the delay and overdue time boundaries.
  We assume the same network as shown by Figure \ref{netw1}, where the UAVs  $drone1$, $drone2$ and  $drone3$  are initially positioned are  Base Stations $B1$, $B2$ and  $B3$, respectively; and all the UAVs are assumed to have the same speed v=800 m/min.
    \FloatBarrier 
\subsubsection{Effect of delay constraints on path design} 

\begin{figure}[ht!]
\begin{floatrow}
\ffigbox[6cm]{%
   \includegraphics[scale=0.4]{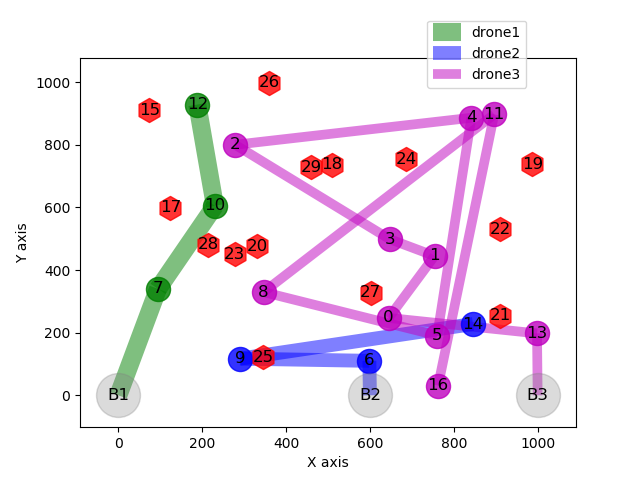} 
}
{%
    \caption{Paths details (waiting=$30$).
 \label{rr1}}
}
\capbtabbox[6cm]{%
\scriptsize
  \begin{tabular}{|p{0.8cm}p{1cm}p{0.7cm}p{1.4cm}|}
 \hline
 UAV name& speed (m/min)& Source & Returning path\\\hline
Drone1  & 800 & B1 & [12, B4]\\
Drone2  & 800 & B2 & [14, 13, B3]\\
Drone3  & 800 & B3 & [16, B3]\\\hline
 \end{tabular}
}{%
     \caption{Data delivery  when UAVs may be late \newline for no more than 30 min.
   \label{tt1}}
}
\end{floatrow}
\end{figure}

 Figure \ref{rr1} and Table \ref{tt1} represent the case where, each sensor node is visited if the arrival of a drone is late for no more than 30 min.

 Figure \ref{rr1} shows that 14 sensor node could not be visited (see red hexagon shaped nodes).  The UAV $drone3$  visit most of visited nodes, whereas other UAVs could visit only 3 sensors each.
Table \ref{tt1} shows that Drone1  delivers to Base Station $B4$ via node $12$, Drone2 to Base Station $B3$  via node $14$ and then $13$ and Drone3 to Base station $B3$ via node $16$.

\begin{figure}[ht!]
\begin{floatrow}
\ffigbox[6cm]{%
   \includegraphics[scale=0.4]{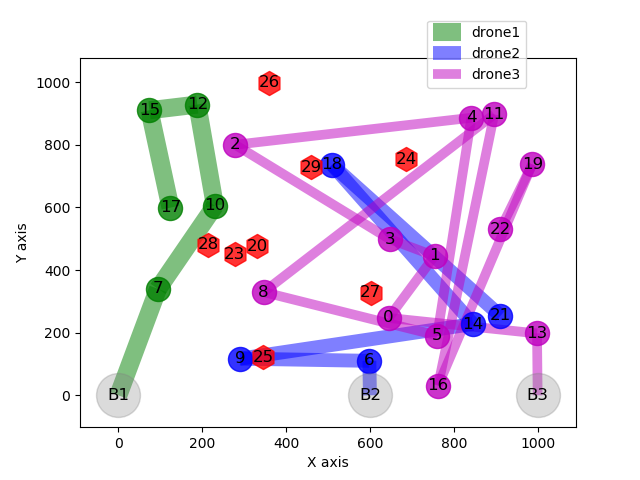}  
}
{%
 \caption{Paths details (waiting=$60$).
 \label{rr2}
}}
\capbtabbox[6cm]{%
\scriptsize
\begin{tabular}{|p{0.8cm}p{1cm}p{0.7cm}p{1.4cm}|}
 \hline
 UAV name& speed (m/min)& Source & Returning path\\\hline
Drone1  & 800 & B1 & [17, B4]\\
Drone2  & 800 & B2 & [21, 13, B3]\\
Drone3  & 800 & B3 & [22, 19, B5]\\\hline
 \end{tabular}
}{%
     \caption{Data delivery  when UAVs may be late for no more than 60 min.}
   \label{TT2}
}
\end{floatrow}
\end{figure}

 Table \ref{TT2}  shows the data delivery paths when UAVs may be late no more than 60 min, and Figure \ref{rr2} shows  the data collection path with this setting.

When changing the late threshold to 60 min, Table \ref{TT2} shows that that the delivery paths changed, and the delivery is done as the last column of the table shows.

Comparing with Figure \ref{rr1},  Figure \ref{rr2} reveals that $drone3$ does not change the path, but the other UAVs extend their path to two more sensor nodes each. This results in 10 sensor nodes to be missed as shown by the  figure.

  \FloatBarrier
 
  %\FloatBarrier

\begin{figure}[ht!]
\begin{floatrow}
\ffigbox[6cm]{%
  \includegraphics[scale=0.4]{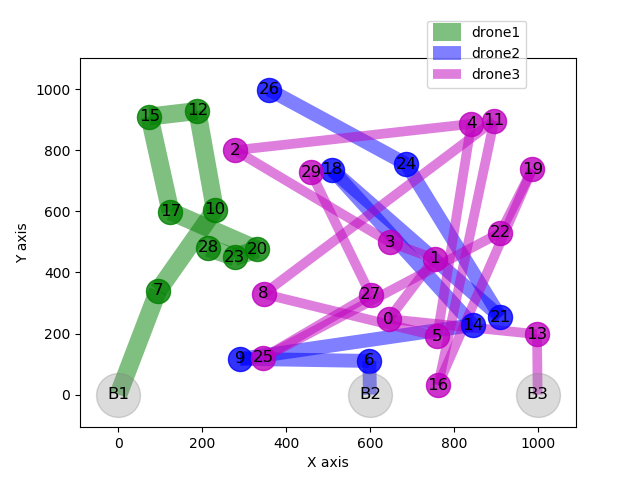}
}
{%
  \caption{Paths details (waiting=$\infty$).
 \label{rr3}}
}
\capbtabbox[6cm]{%
\scriptsize
\begin{tabular}{|p{0.8cm}p{1cm}p{0.7cm}p{1.4cm}|}
 \hline
 UAV name& speed (m/min)& Source & Returning path\\\hline
Drone1  & 800 & B1 & [28, 7, B1]\\
Drone2  & 800 & B2 & [26, 12, B4]\\
Drone3  & 800 & B3 & [29, 15, B4]\\\hline
 \end{tabular}
}{%
     \caption{Data delivery  when UAVs may be late for no more than 60 min.
   \label{tt3}}
}
\end{floatrow}
\end{figure}

  Table \ref{tt3} and Figure \ref{rr3} respectively show the data delivery and collection paths, when UAVs may be late for a very long time.

  %\FloatBarrier

 Comparing with the previous two cases, Table \ref{tt3} shows that the delivery paths keep on changing different.

Figure \ref{rr3} shows that all nodes have been visited and the path of each drone has been extended, to cover more nodes.

\begin{figure}[ht!]
 \centering
 \includegraphics[scale=0.4]{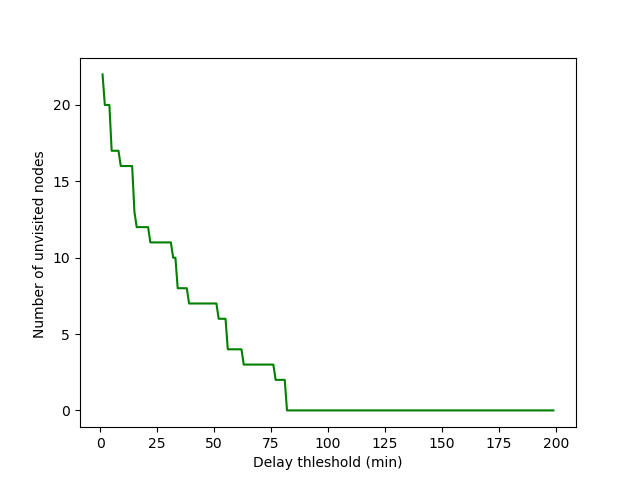} 
  \caption{Number of unvisited nodes.
 \label{rr4}}
 \end{figure}

 \FloatBarrier
Figure \ref{rr4} reveals the changes in the number of missed nodes while the late threshold evolves. The figure shows that when the delay threshold increases, the number of unvisited nodes remains constant or decreases (it does never increase), until it converges to zero. This is justified by the fact that, allowing a longer delay increases the chance for a node to be visited.

\subsubsection{Effect of overdue constraints on path design} 

In this section, we study  the effect of the waiting constraints. Here, sensors can only be visited after some fixed time called the waiting threshold.

\begin{figure}[ht!]
    \centering
    \begin{subfigure}[b]{0.48\textwidth}
        \includegraphics[scale=0.4]{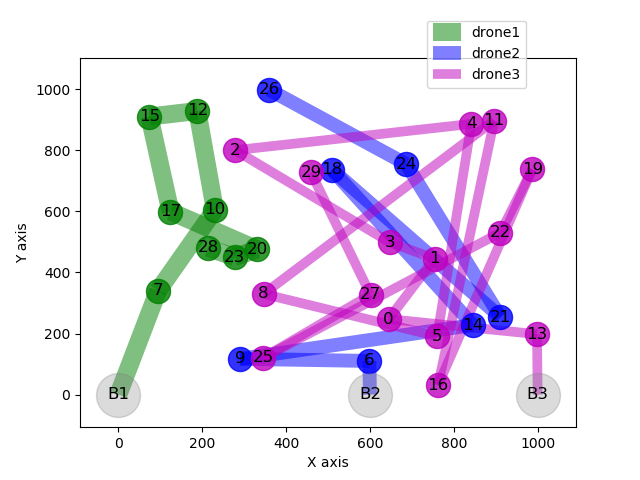} 
         \caption{Waiting for 0.0 min
        \label{rrr1}}
    \end{subfigure}
    ~ %add desired spacing between images, e. g. ~, \quad, \qquad, \hfill etc. 
    \begin{subfigure}[b]{0.48\textwidth}
         \includegraphics[scale=0.4]{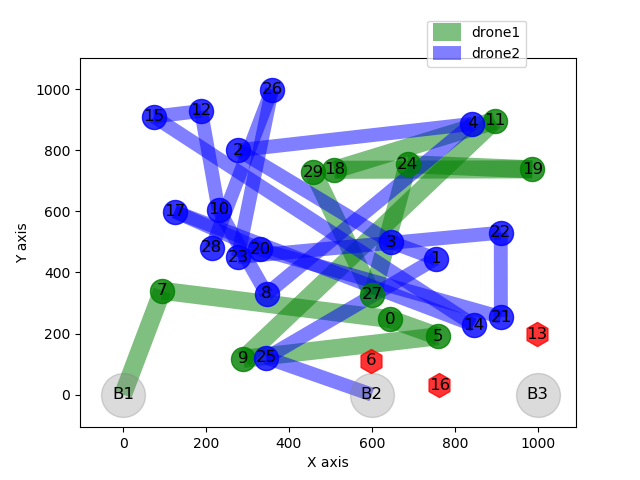} 
             \caption{Waiting for 0.3  min
        \label{rrr2}}
    \end{subfigure}

    \begin{subfigure}[b]{0.48\textwidth}
         \includegraphics[scale=0.4]{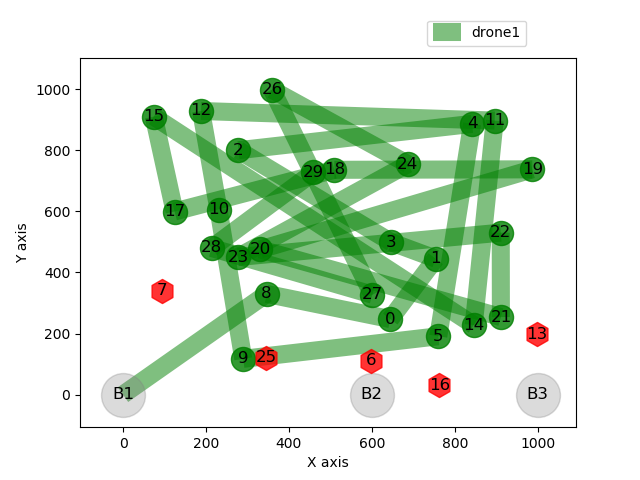} 
            \caption{Waiting for 0.5  min.
        \label{rrr3}}
    \end{subfigure}
        \begin{subfigure}[b]{0.48\textwidth}
         \includegraphics[scale=0.4]{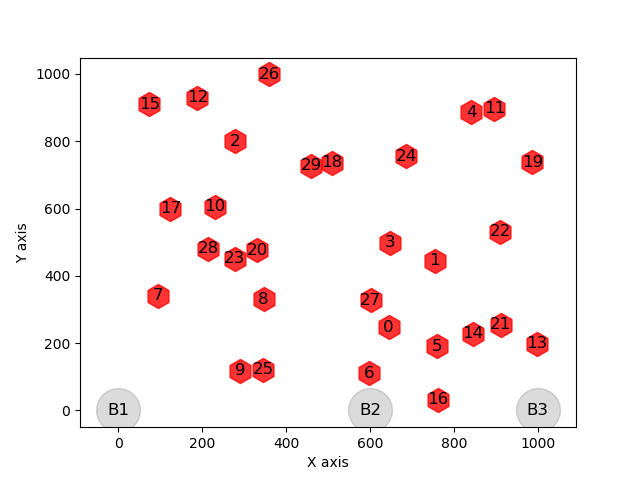} 
           \caption{Waiting for 0.7  min.
        \label{rrr4}}
    \end{subfigure}
     \caption{Visitation constrained by the waiting time.
    \label{w}}
\end{figure}

 Figure \ref{w} shows how paths corresponding to different thresholds are generated. Figure \ref{rrr1} considers a case where there is no waiting limitation (waiting time is zero) and clearly all sensors are visited. When the waiting time is set to 0.3 minutes (18 seconds) in Figure \ref{rrr2} three nodes are not visited and the  Drone 3 could not visit any sensor. Setting the waiting threshold to 0.5 min, only  UAV Drone 1 could visit and 5 sensors were missed.
 Setting the waiting time to 0.7 min, no node could be visited.  This is because there is a specific time for UAVs to arrive at each sensors which depend on the distance and speed. If the distance is small and the UAV can arrive earlier  than the waiting threshold, the visitation is impossible.

 % \FloatBarrier

   %\FloatBarrier
 \begin{table}[h!]

 \centering
  \scalebox{0.7}{
\begin{subtable}[t]{0.77\textwidth}
       \begin{tabular}{|p{1.4cm}p{1.8cm}p{1.5cm}p{1.8cm}|}
 \hline
 UAV name& speed (m/min)& Source & Returning path\\\hline
Drone1  & 800 & B1 & [28, 7, B1]\\
Drone2  & 800 & B2 & [26, 12, B4]\\
Drone3  & 800 & B3 & [29, 15, B4]\\\hline
 \end{tabular}
         \caption{Data delivery when the waiting   threshold is 0 min.
        \label{ttt1}}
     \end{subtable}
\begin{subtable}[t]{0.77\textwidth}
       \begin{tabular}{|p{1.4cm}p{1.8cm}p{1.5cm}p{1.8cm}|}
 \hline
 UAV name& speed (m/min)& Source & Returning path\\\hline
Drone1  & 800 & B1 & [29, 12, B4]\\
Drone2  & 800 & B2 & [28, 7, B1]\\
Drone3  & 800 & B3 & [B3]\\\hline
 \end{tabular}
         \caption{Data delivery when the waiting    threshold is 0.3 min.
        \label{ttt1}}
    \end{subtable}}
  \caption{Data delivery when the waiting threshold is small.}  
  \label{ddl}
\end{table}

 \begin{table}[h!]

 \centering
  \scalebox{0.7}{
\begin{subtable}[t]{0.77\textwidth}
       \begin{tabular}{|p{1.4cm}p{1.8cm}p{1.5cm}p{1.8cm}|}
 \hline
 UAV name& speed (m/min)& Source & Returning path\\\hline
Drone1  & 800 & B1 & [29, 12, B4]\\
Drone2  & 800 & B2 & [ B2]\\
Drone3  & 800 & B3 & [ B3]\\\hline
 \end{tabular}
        \caption{Data delivery when the waiting   threshold is 0.5 min.
        \label{ttt3}}
    \end{subtable}
    
\begin{subtable}[t]{0.77\textwidth}
          \begin{tabular}{|p{1.2cm}p{1.8cm}p{1.5cm}p{1.8cm}|}
 \hline
 UAV name& speed (m/min)& Source & Returning path\\\hline
Drone1  & 800 & B1 & [B1]\\
Drone2  & 800 & B2 & [ B2]\\
Drone3  & 800 & B3 & [ B3]\\\hline
 \end{tabular}
          \caption{Data delivery when the waiting  threshold is 0.7 min.
        \label{ttt4}}
    \end{subtable}}
    \caption{Data delivery when the waiting threshold is high.}
    \label{ddh}
    \end{table}

Tables \ref{ddh} and \ref{ddl}, show deliveries corresponding to visitations shown in Figure  \ref{w}. Note that if a UAV does not visit any sensor, it remains at its initial position.

\begin{figure}[ht!]
 \centering
 \includegraphics[scale=0.4]{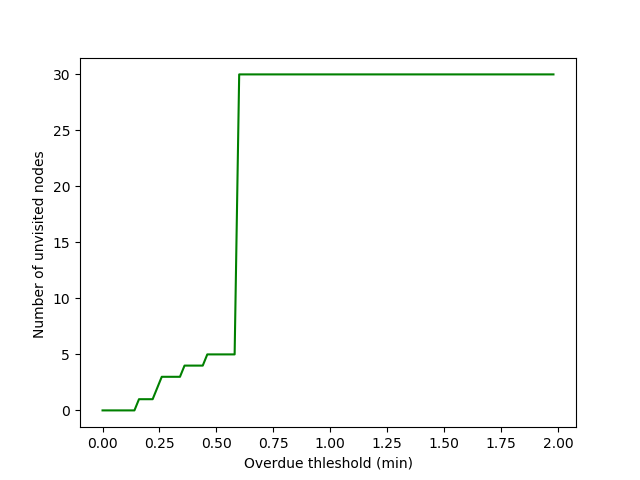} 
  \caption{Number of unvisited nodes.
 \label{rrrr3}}
 \end{figure}

Keeping on changing the waiting threshold, Figure \ref{rrrr3} shows the corresponding number of missed sensors.

 %\FloatBarrier

 Figure \ref{rrrr3}  reveals that the number of missed nodes increases as the waiting threshold increases and converges to the total number of sensors to be visited.
\subsubsection{Persistent visitation analysis} 

In this subsection, we study the trend of  paths  while UAVs  persistently visit sensors. Persistent visitation is done by resetting the initial base stations for the next visit, to the destination of the previous visitation.

 %\FloatBarrier 

\begin{figure}[ht!]
    \centering
    \begin{subfigure}[b]{0.48\textwidth}
        \includegraphics[scale=0.4]{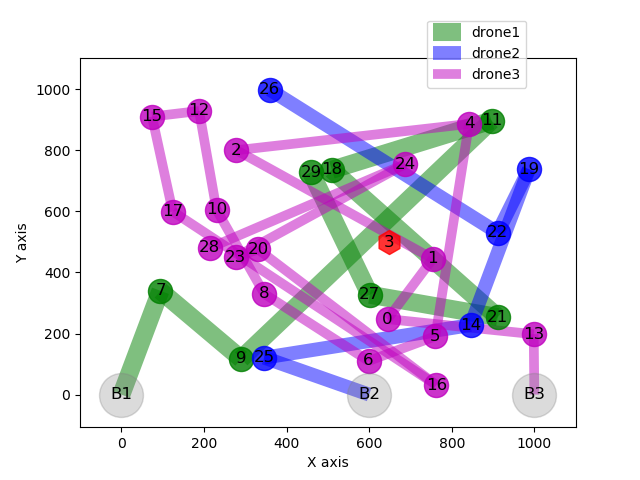} 
         \caption{First visitation.
        \label{v1}}
    \end{subfigure}
    ~ %add desired spacing between images, e. g. ~, \quad, \qquad, \hfill etc. 
    \begin{subfigure}[b]{0.48\textwidth}
         \includegraphics[scale=0.4]{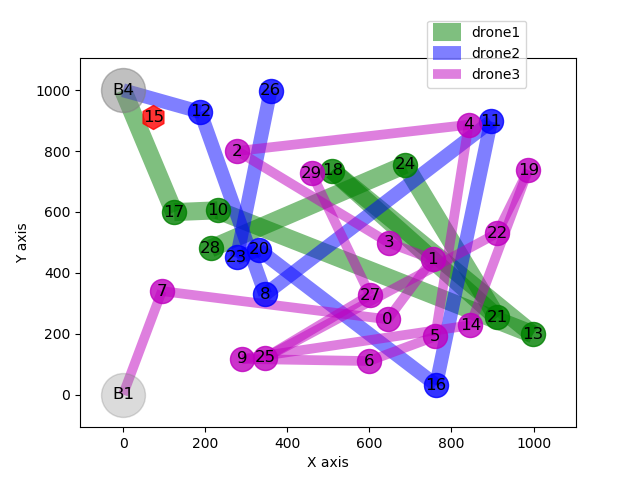} 
         \caption{Second visitation.
        \label{v2}}
    \end{subfigure}
    \caption{First two visitations.}
    \end{figure}

\begin{figure}[ht!]
    \begin{subfigure}[b]{0.48\textwidth}
         \includegraphics[scale=0.4]{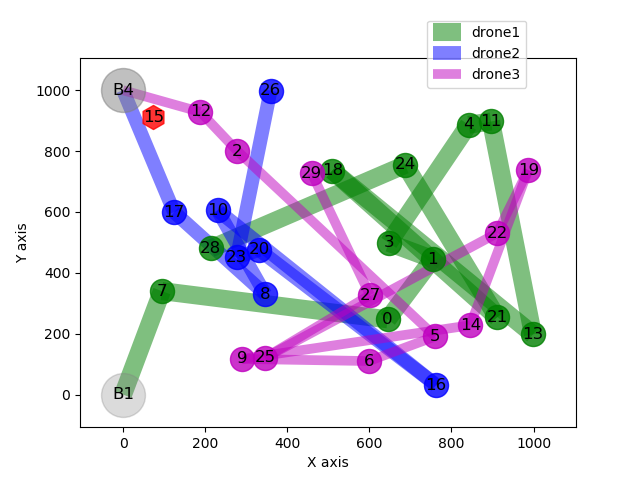} 
         \caption{Third visitation.
        \label{v3}}
    \end{subfigure}
        \begin{subfigure}[b]{0.48\textwidth}
         \includegraphics[scale=0.4]{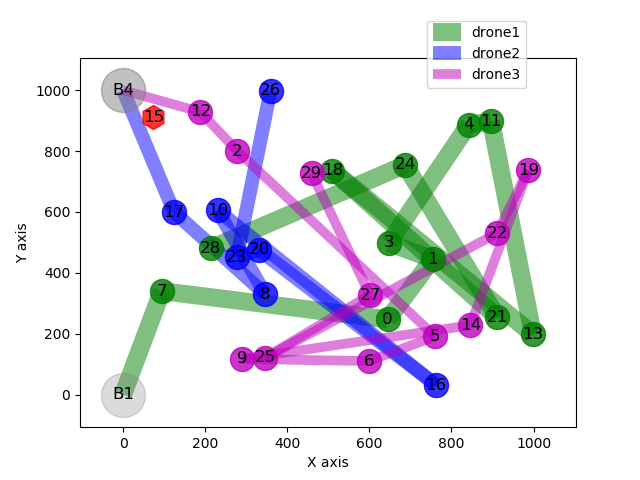} 
         \caption{Fourth visitation.
        \label{v4}}
    \end{subfigure}
     \caption{Next two visitations.}
    \label{v}
\end{figure}
  %\FloatBarrier

Figure \ref{v} shows four consecutive visitations and deliveries when the waiting threshold is set to 12 seconds. Figure \ref{v1} shows that only node $3$ has not been visited. Figure \ref{v2} shows that a new node (node $15$) has not been visited  and all visitation paths change. Figure \ref{v3} shows that paths keep on changing and non-visited nodes remain the same as the previous visitation.

Note that Figure \ref{v4} is exactly the same as Figure \ref{v3}. This shows that all the following paths will be the same as Figure \ref{v3} and hence paths generation may converge to specific paths. This is a special case where UAVs' initial positions become the same as their optimal destination.

We now consider the constraint free visitations and study the  pattern of generated paths. 
\begin{figure}[ht!]
    \begin{subfigure}[b]{0.48\textwidth}
    \centering
        \includegraphics[scale=0.38]{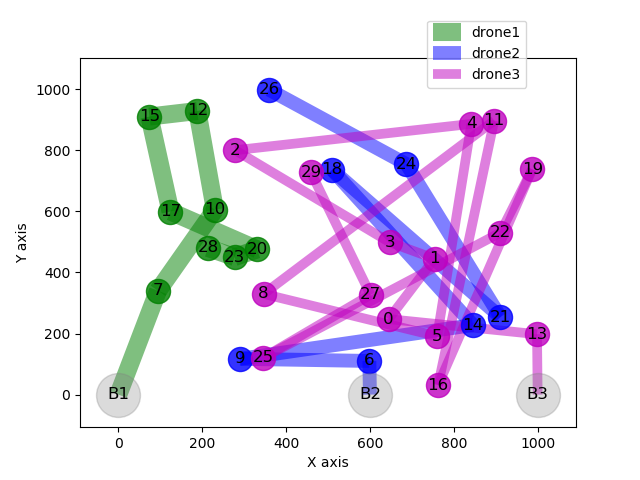} 
         \caption{First visitation
        \label{c1}}
     \end{subfigure}
    ~ %add desired spacing between images, e. g. ~, \quad, \qquad, \hfill etc. 
           \begin{subfigure}[b]{0.48\textwidth}
       \centering
         \includegraphics[scale=0.38]{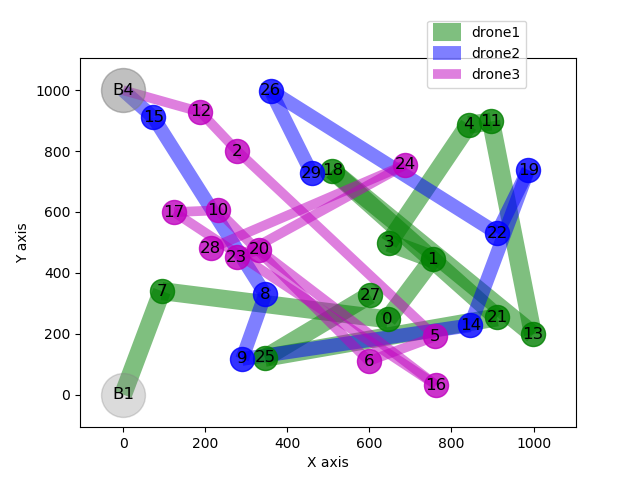} 
         \caption{Second visitation.
        \label{c2}}
            \end{subfigure}

    \begin{subfigure}[b]{0.48\textwidth}
\centering
         \includegraphics[scale=0.38]{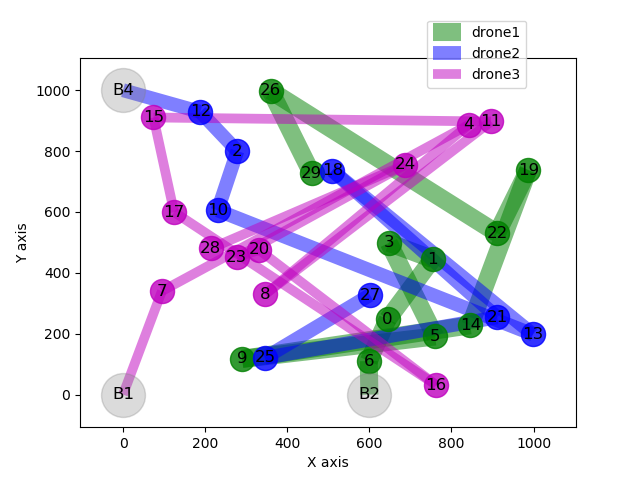} 
         \caption{Third visitation.
        \label{c3}}
    \end{subfigure}
          \begin{subfigure}[b]{0.48\textwidth} 
       \centering
         \includegraphics[scale=0.38]{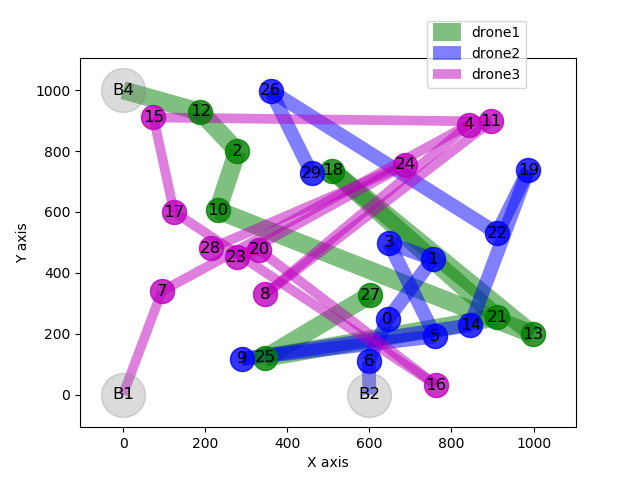} 
         \caption{Fourth visitation.
        \label{c4}}
        \end{subfigure}
        
    \begin{subfigure}[b]{0.48\textwidth}
  \centering
        \includegraphics[scale=0.38]{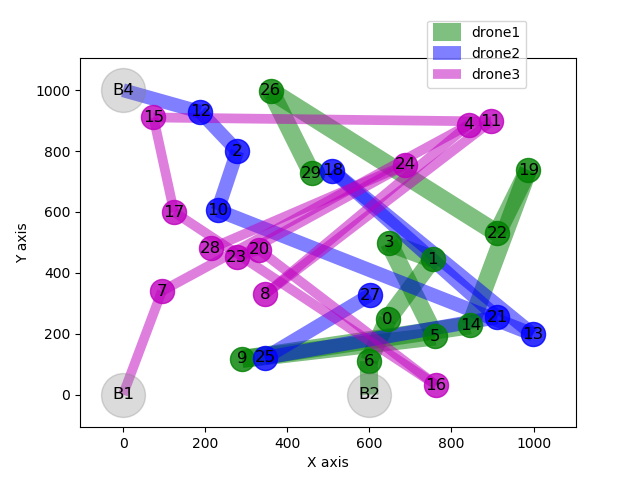} 
         \caption{sixth visitation.
        \label{c5}}
    \end{subfigure} 
    \begin{subfigure}[b]{0.48\textwidth}
    \centering
         \includegraphics[scale=0.38]{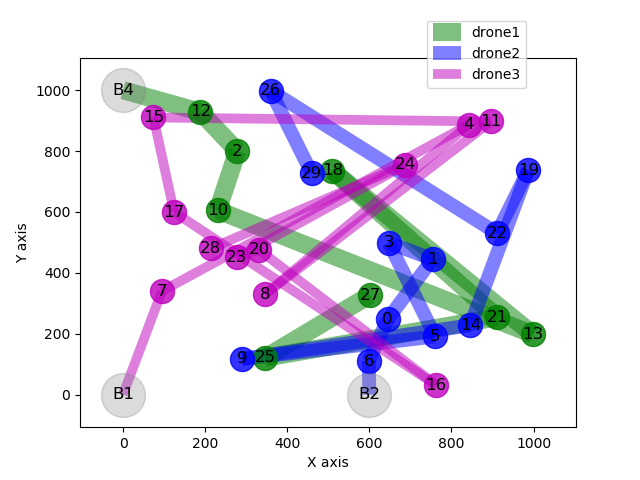} 
         \caption{seventh visitation.
        \label{c6}}
    \end{subfigure}
%
%     {Unconstrained visitations}
%    \label{c}
\end{figure}

 %\FloatBarrier
 
Figures \ref{c1}, \ref{c2}, \ref{c3}, \ref{c4}, \ref{c5}, \ref{c6} show the first 6 consecutive unconstrained  visitations. 
The path change and from the third visitation,  paths generation becomes periodic: third visitation is same as the fifth, and the fourth visitation is the same as the sixth. This shows that after each next visitations paths generation remains the same. This is justified by the fact that from each base station, each UAV  has a single optimal destination.

\FloatBarrier
 
\section{Conclusion and future work}\label{sec5}
In this paper, a model for data muling targeting revisit costs minimization for a team of UAVs has been provided. A mathematical formulation of the model 
has been presented and the underlying problem has been polynomially reduced to another NP-hard problem and hence proved to be an NP-hard problem. A heuristic solution to  the problem has been provided and its performance evaluated through simulation experiments. Simulation results have revealed different path distribution patterns under different experimental settings and the impact of these settings on path length fairness and related energy cost. 
 Furthermore, simulations show that consecutive paths generation becomes periodic after some number of visitations. While this paper has presented the basis of a data muling model aiming at supplementing traditional traffic engineering techniques used in sink networks, several network and traffic aspects related to the proposed data muling model still need to be investigated. These include the design of efficient communication models that consider the outdoor characteristics of drone-to-sink communication as suggested in \citep{experimental}. Taking advantage of the emerging white space frequency bands as discussed in \citep{whitespace} to achieve drone-to-sink and drone-to-drone communication is another direction for further work.

\section{Appendix}

\begin{figure}[ht!]
 \begin{subfigure}[b]{0.48\textwidth}
\centering
\includegraphics[scale=0.3]{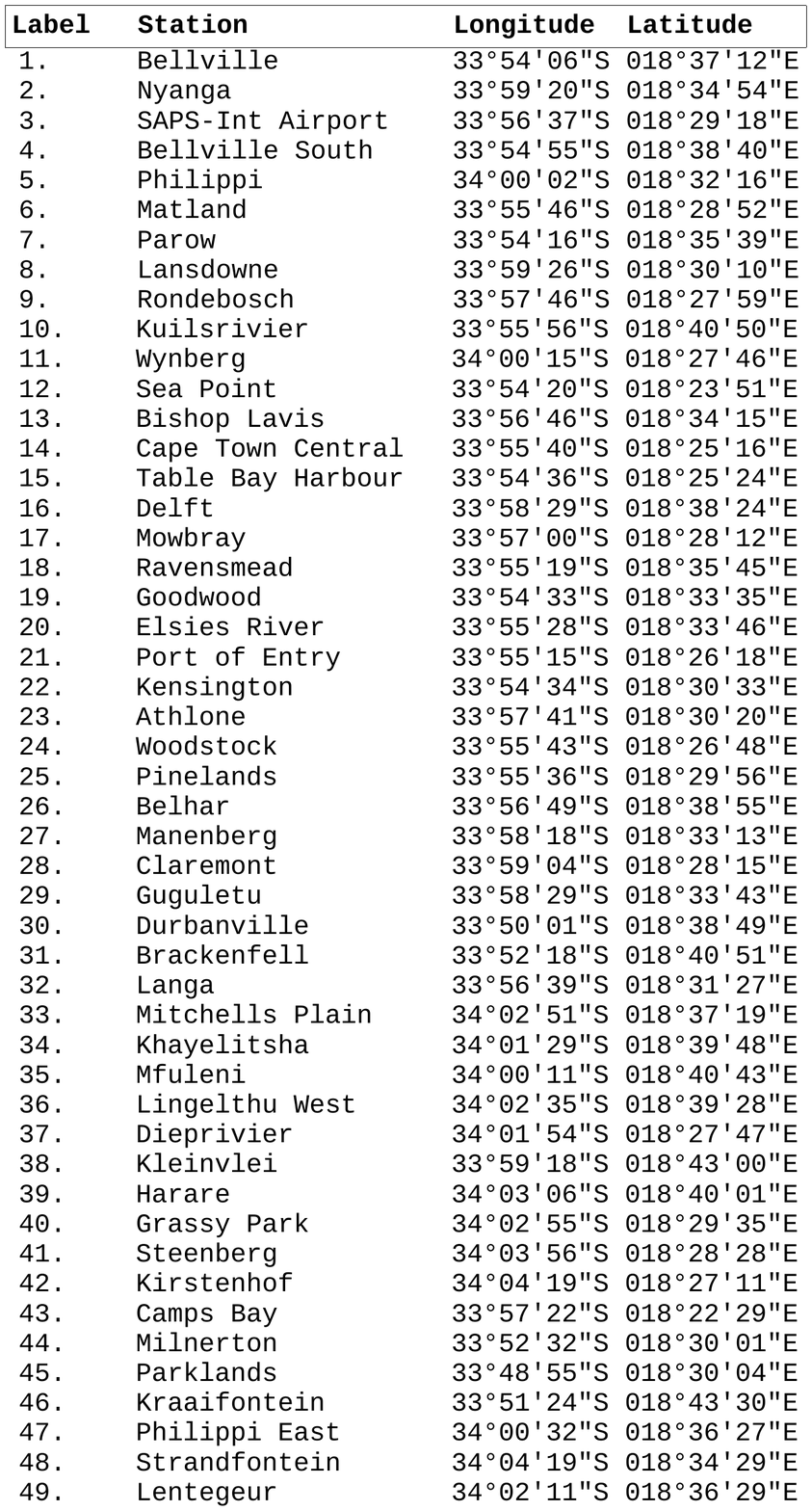}   
 \caption{GPS positions of Cape Town police stations.}
\label{net00}
\end{subfigure}
\begin{subfigure}[b]{0.48\textwidth}
\includegraphics[scale=0.25]{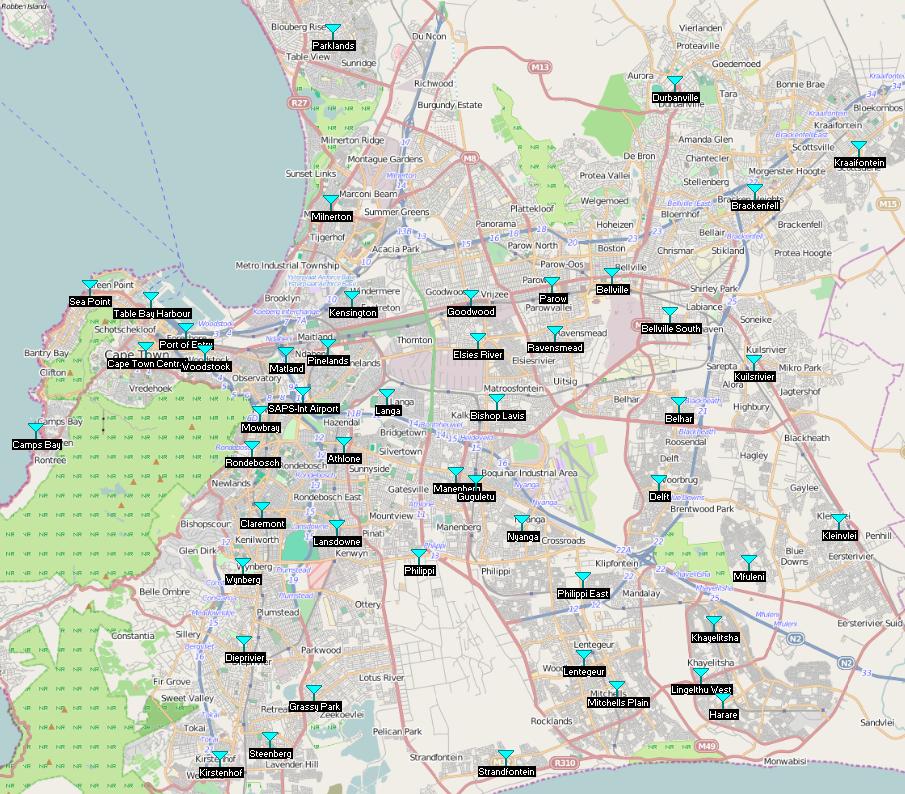}   
 \caption{Positions on the globe's map.
\label{net10}}
\end{subfigure}
 \caption{Raw real network study.}
\end{figure}

We  assume  the public safety network consisting of Cape Town (South Africa) police stations as collection points of a ground sensor network used for example for city safety or traffic control. Cape Town police stations are labelled in terms of integers in interval $[1,49]$ and their GPS coordinates are  used as their positions  (see Figure \ref{net00}). The corresponding positions on a map are shown in Figure \ref{net10}.

 \FloatBarrier

 \FloatBarrier